\documentclass[authoryear,final,review,12pt]{elsarticle}
\usepackage[lmargin=1.1in, rmargin=1.1in, tmargin=1in, bmargin=1in, nohead]{geometry}

\usepackage{multirow}

\usepackage{amssymb,amsmath}
\usepackage{amsthm}
\usepackage{rotating}

\usepackage{epsfig}
\usepackage{epsf}
\usepackage{psfrag}
\usepackage{caption}
\usepackage{latexsym, graphics, psfrag, amscd, amssymb, pb-diagram}
\usepackage[bottom]{footmisc}

\usepackage{color}
\usepackage[all,cmtip]{xy}

\usepackage[lined, ruled, linesnumbered]{algorithm2e}
\usepackage{changepage}
\usepackage{xassoccnt}
\usepackage{lastpage}
\usepackage{hyperref}
\usepackage{float}
\usepackage{setspace}

\newcommand{\myvec}[1]%
{\stackrel{\raisebox{-2pt}[0pt][0pt]{\small$\rightharpoonup$}}{#1}}

\NewTotalDocumentCounter{totalfigures}
\NewTotalDocumentCounter{totaltables}
\NewTotalDocumentCounter{appendixchapters}
\DeclareAssociatedCounters{figure}{totalfigures}
\DeclareAssociatedCounters{table}{totaltables}
\newcommand{\totalpages}{\zref@extractdefault{LastPage}{page}{0}}

\usepackage{lipsum}
\makeatletter
\def\ps@pprintTitle{%
 \let\@oddhead\@empty
 \let\@evenhead\@empty
 \def\@oddfoot{}%
 \let\@evenfoot\@oddfoot}
\makeatother

\begin{document}
\singlespacing

\begin{frontmatter}

\title{Developing Univariate Neurodegeneration Biomarkers with Low-Rank and Sparse Subspace Decomposition}

\author[Ludong]{\mbox{Gang Wang}\corref{cor1}}
\author[asu]{Qunxi Dong}
\author[asu]{Jianfeng Wu}
\author[Banner]{Yi Su}
\author[Banner]{Kewei Chen}
\author[Ludong1]{Qingtang Su}
\author[Ludong1]{\\Xiaofeng Zhang}
\author[Ludong1]{Jinguang Hao}
\author[Ludong1]{Tao Yao}
\author[Ludong1]{Li Liu}
\author[SDUFE]{\\Caiming Zhang}
\author[Mayo]{Richard J Caselli}
\author[Banner]{Eric M Reiman}
\author[asu]{\mbox{Yalin Wang}\corref{cor1}}
\author{\mbox{for the Alzheimer's Disease Neuroimaging Initiative}}
\address[Ludong]  {Ulsan Ship and Ocean College,\\
Ludong University,
Yantai, China}

\address[asu] {School of Computing, Informatics, and Decision Systems Engineering,\\
Arizona State University,
Tempe, AZ, USA}
\address[Banner]  {Banner Alzheimer's Institute and Banner Good Samaritan Pet Center \\
Phoenix, AZ, USA}
\address[Ludong1]  {School of Information and Electrical Engineering, \\
Ludong University,
Yantai, China}
\address[SDUFE]  {Shandong Province Key Lab of Digital Media Technology \\
Shandong University of Finance and Economics,
Jinan, China}\address[Mayo]  {Department of Neurology \\
Mayo Clinic Arizona Scottsdale, AZ, USA}




\begin{abstract}
\let\thefootnote\relax\footnotetext{Data used in preparation of this article were obtained from the Alzheimer\textquoteright s Disease Neuroimaging Initiative (ADNI) database (adni.loni.usc.edu). As such, the investigators within the ADNI contributed to the design and implementation of ADNI and/or provided data but did not participate in analysis or writing of this report. A complete listing of ADNI investigators can be found at: https://adni.loni.usc.edu/wp-content/uploads/how\textunderscore to\textunderscore apply/ADNI\textunderscore Acknowledgement\textunderscore List.pdf \\
\textbf{E-mail:} ylwang@asu.edu, gangwang1970@ldu.edu.cn}\noindent Cognitive decline due to Alzheimer's disease (AD) is closely associated with brain structure alterations captured by structural magnetic resonance imaging (sMRI). It supports the validity to develop sMRI-based univariate neurodegeneration biomarkers (UNB). However, existing UNB work either fails to model large group variances or does not capture AD dementia (ADD) induced changes. We propose a novel low-rank and sparse subspace decomposition method capable of stably quantifying the morphological changes induced by ADD. Specifically, we propose a numerically efficient rank minimization mechanism to extract group common structure and impose regularization constraints to encode the original 3D morphometry connectivity. Further, we generate regions-of-interest (ROI) with group difference study between common subspaces of $A\beta+$ AD and $A\beta-$ cognitively unimpaired (CU) groups. A univariate morphometry index (UMI) is constructed from these ROIs by summarizing individual morphological characteristics weighted by normalized difference between $A\beta+$ AD and $A\beta-$ CU groups. We use hippocampal surface radial distance feature to compute the UMIs and validate our work in the Alzheimer's Disease Neuroimaging Initiative (ADNI) cohort. With hippocampal UMIs, the estimated minimum sample sizes needed to detect a 25$\%$ reduction in the mean annual change with 80$\%$ power and two-tailed $P=0.05$ are 116, 279 and 387 for the longitudinal $A\beta+$ AD, $A\beta+$ mild cognitive impairment (MCI) and $A\beta+$ CU groups, respectively. Additionally, for MCI patients, UMIs well correlate with hazard ratio of conversion to AD ($4.3$, $95\%$ CI=$2.3-8.2$) within 18 months. Our experimental results outperform traditional hippocampal volume measures and suggest the application of UMI as a potential UNB.

\end{abstract}
\begin{keyword}
Magnetic resonance imaging (MRI), Alzheimer\textquoteright s disease, Subspace decomposition, Univariate morphometry index, Minimum sample size, Cox proportional hazard model
\end{keyword}
\end{frontmatter}

\section{Introduction}\label{sec:1}
Alzheimer's disease (AD) is a progressive neurodegenerative disease and also the most common type of dementia. Its prevalence is predicted to triple to 13.8 million by 2050. It is generally agreed that effective diagnosis and treatment of AD could have enormous public health benefits. Prior research on structural magnetic resonance imaging (sMRI) supported its validity as a potential noninvasive neurodegeneration biomarker of AD \citep{Frisoni:NatRev10,Sperling:PreAD11,Jack:ATN16,Zhao:NIMG17}. Such sMRI-based imaging biomarkers will facilitate the development of effective interventions that may postpone (or prevent) symptomatic AD~\citep{caselli:AD13,langbaum:Natrev13}.

Structural MRI-based measures of atrophy including whole brain~\citep{Fox:neu99,chen:psychiatry07,Stonnington:Predicting}, entorhinal cortex~\citep{cardenas:neuaging11}, hippocampus~\citep{jack:neurology03,Thompson:nimg04,denheijer:brain10,wang:nimg11,Dong:NICL19} and temporal lobe volumes~\citep{hua:nimg10}, as well as ventricular enlargement~\citep{jack:neurology03,Thompson:nimg04,wang:nimg11,dong2020applying}, correlate closely with changes in cognitive performance, inspiring various neuroimaging biomarker approaches. Among them, hippocampal atrophy measures from sMRI are widely used, as hippocampal morphometry changes are apparent in the early stages of memory decline and may anticipate progression to mild cognitive impairment (MCI) and AD~\citep{jack:neurology03}.

While much AD imaging biomarker research has been devoted to group difference-based analyses, a univariate neurodegeneration biomarker (UNB) based on an individual patient's brain scans with high diagnostic accuracy would be highly desirable for clinical use~\citep{Sabuncu:PNAS16}. Such a personalized measure may overcome inflated Type I error due to multivariate comparisons. For example, for randomized clinical trials (RCT), regulatory agencies, including the Food and Drug Administration (FDA), requires conventional univariate hypothesis testing and its associated statistical power analysis~\citep{langbaum:Natrev13}. Meanwhile, a single MRI-based measure of cerebral atrophy was used as a neurodegeneration biomarker in the recently proposed AD amyloid/tau/neurodegeneration (A/T/N) framework~\citep{Jack:ATN16}. It is a descriptive system for categorizing multidomain biomarker findings at the individualized level in a format that is easy to understand and use. Rather than conceptualizing AD primarily as a clinicopathological entity, in this system AD is diagnosed using biomarkers such as brain imaging or by measuring substances in the cerebral spinal fluid. This change may allow preclinical AD diagnosis on presymptomatic patients~\citep{Knopman:AD18}. However, a recent work~\citep{Illan:AD18} reported that the currently available neurodegeneration biomarkers, including hippocampal volume and cortical signature of AD~\citep{Dickerson:CerCor09}, were poorly correlated both in their whole dataset and along the AD continuum. Therefore, to advance computational neuroanatomy to clinical usage, developing a robust method to quantify brain sMRI difference using a statistically powerful UNB will be highly advantageous for clinical diagnosis and prognosis.

In the past decade, a variety of univariate morphological MRI biomarker algorithms have been developed~\citep[e.g.][]{Racine:AD18,cardenas:neuaging11,Cortechslab,Vemuri:NIMG08,Tu:NEIN20}. Meanwhile, a few multivariate analysis frameworks~\citep{Hua:NIMG11,Gutman:NIMG13} took statistical or machine learning approaches to optimize the minimum sample size estimation for clinical trials. Among them, most are based on regions-of-interest (ROI) analyses and obtain a UNB by summarizing image information from these ROIs. Such methods enjoy good biological interpretability and excellent computational efficiency. However, there are a number of issues: (1) In their modeling, such approaches do not take into account the strong noise and the large within-group variance on the obtained sMR images; (2) When setting up training data, they usually use the clinical symptom-based AD diagnosis without the confirmation of imaging or fluid $A\beta$ information. These difficulties have largely limited the statistical power and generalibility of their UNB due to excessive imaging noise, strong individual morphological differences and the inhomogeneity of dementia.

To address these challenges, it is important to generate reliable and robust sMRI ROIs that reflect intrinsic structural differences caused by AD dementia (ADD). Recently, research on low-rank and sparse subspace decomposition by rank minimization and sparse constraint has attracted strong interest~\citep{Cao:Neural17,Jang:CVPR2016, Javed:CSVT16}. In particular, robust principal component analysis (RPCA)~\citep{Candes:JACM11} through principal component pursuit (PCP) decomposes a data matrix into two subspaces which include a low-rank component and a sparse component. The low-rank subspace may mine the common structure belonging to a specific group, and the sparse component can detect salient features belonging to a specific individual~\citep{Hastie:Sparsity15}. When individual brain image measurements of the same group subjects are stacked into columns to generate the group observation matrix, we may expect its low-rank components correspond to the group common structure. Except for this group common structure, individual image differences correspond to the sparse component. Furthermore, one of the hallmarks of AD is the accumulation of beta-amyloid plaques ($A\beta$) in human brains and a positive $A\beta$ reading is now accepted as 
`dementia due to AD' together with the presence of clinical symptoms. In this paper, we propose a novel subspace decomposition-based method to generate the ROIs based on the group difference from extracted low-rank components of $A\beta$ positive AD and $A\beta$ negative cognitively unimpaired (CU) groups and later define a univariate neurodegeneration biomarker, univariate morphometry index (UMI), which may reflect intrinsic morphological changes induced by ADD and also have strong generalization ability on new subjects. Previous research~\citep[e.g.][]{Thompson:nimg04,Apostolova:NA10,Qiu:TIP10,Costafreda:NIMG11,Dong:NICL19} has demonstrated that surface-based hippocampal morphometry analysis offers advantages over volume measures. Therefore, we apply our subspace decomposition method to hippocampal measures obtained with our prior hippocampal morphometry study~\citep{shi:nimg13} to construct UMIs. We hypothesize that our low-rank and sparse subspace decomposition framework together with surface-based measures may be used to compute robust and efficient UNBs and improve their statistical power compared with the traditional univariate biomarker, such as hippocampal volumes~\citep{jack:neurology03,Cortechslab}. Here, we set out to validate our hypothesis on brain sMR images from the Alzheimer's Disease Neuroimaging Initiative (ADNI) cohort.

\section{Material and Methods}\label{sec:2}

\subsection{Subjects}

Data is downloaded from the Alzheimer's Disease Neuroimaging Initiative (ADNI) database (\citep{mueller:ncna05}, adni.loni.usc.edu). ADNI is the result of efforts of many co-investigators from a broad range of academic institutions and private corporations. Subjects have been recruited from over 50 sites across the U.S. and Canada. The primary goal of ADNI is to test whether biological markers, such as serial MRI and positron emission tomography (PET), combined with clinical and neuropsychological assessments, can measure the progression of MCI and early AD. Subjects originally recruited for ADNI-1 and ADNI-GO had the option to be followed in ADNI-2. For up-to-date information, see www.adniinfo.org.

In this study, we use brain structural MRI data from $151$ A$\beta$ positive ($A\beta+$) AD patients and $271$ A$\beta$ negative ($A\beta-$) CU to identify the ROIs which are further used to define UMI. To validate our algorithm, we study (1) longitudinal brain imaging changes via group difference and minimum sample size estimation; and (2) future clinical decline prediction for MCI patients via survival analysis. In the first experiment, we use structural MRI data from $165$ longitudinal $A\beta+$ subjects, including $46$ AD, $55$ MCI and $64$ CU subjects. All subjects of each longitudinal group underwent two tests, including the baseline test and a 24-months test. In the second experiment, we use $155$ MCI patients including $59$ MCI subjects who have converted to probable AD and $96$ non-converter MCI subjects within 18-months period.

\subsection{System Pipeline Overview}\label{sec:pipeline}

\begin{figure}[htp]
\begin{center}
\includegraphics[width=0.75\textwidth]{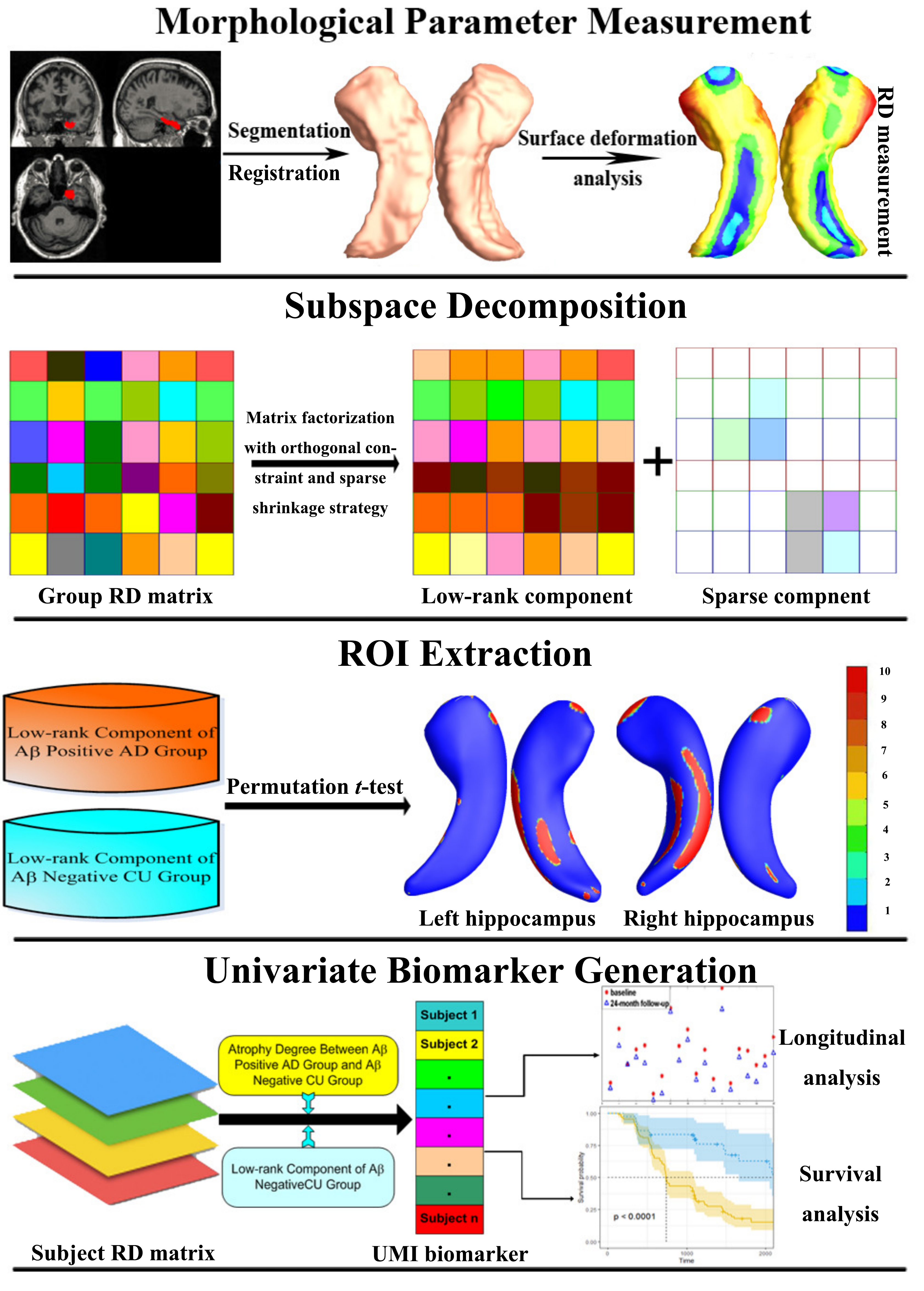}

\caption{The proposed univariate morphometry index (UMI) generation system pipeline, illustrated by the intermediate results.}
\label{fig:pipeline}
\end{center}

\end{figure}

Here we briefly overview the processing pipeline in our univariate morphometry index (UMI) system. Following sections are detailed explanations of each step. The pipeline code is publicly available at https://gsl.lab.asu.edu/software/univariate-neurodegeneration\\-biomarkers.

Fig.~\ref{fig:pipeline} summarizes the overall sequence of steps in the system. First, we reconstruct hippocampal surfaces after we segment hippocampi from T1-weighted MR images. We register hippocampal surfaces across subjects and compute surface morphometry features, radial distance (RD)~\citep{Pizer:TMI99,Styner:PNAS06,Thompson:nimg04} (the first row, Fig.~\ref{fig:pipeline}). Second, we build group morphometry (i.e., radial distance - RD) observation matrices, i.e., the observation matrix of A$\beta$ positive ($A\beta+$) AD group and the observation matrix of A$\beta$ negative ($A\beta-$) CU group, where each individual RD features are concatenated as a column vector. Due to the morphological similarity within the same group, we further decompose each group RD matrix into low-rank component and sparse component based on the matrix factorization with orthogonal constraint and sparsity shrinkage strategy (the second row, Fig.~\ref{fig:pipeline}). Third, with group difference study, we identify significantly different areas between the low rank components of two groups, $A\beta+$ AD group and $A\beta-$ CU group, as the regions-of-interest (ROI) and use them as the significant morphological change regions induced by ADD (the third row, Fig.~\ref{fig:pipeline}). Finally, based on the registration result of an individual hippocampal surface, the UMI of the individual subject is computed through the voxel-wise AD atrophy degrees and the individual atrophy degrees defined on the ROIs (Sec.~\ref{sec:MSSI}). As applications of the computed UMIs, we apply them to three different studies, including longitudinal analysis, sample size estimation, and progression rate predictions from  MCI to probable AD (the fourth row, Fig.~\ref{fig:pipeline})

\subsection{Image Acquisition and Hippocampal Morphometry Surface Feature Computation}~\label{sec:preprocessing}
High-resolution brain structural MRI scans are acquired at multiple ADNI sites using 1.5 Tesla MRI scanners manufactured by General Electric Healthcare, Siemens Medical Solutions, and Philips Medical Systems. For each subject, the T1-weighted MRI scan is collected with a sagittal 3D MP-RAGE sequence. Typical $1.5$T acquisition parameters are repetition time (TR) of $2,400$ ms, minimum full excitation time (TE), inversion time (TI) of $1,000$ ms, flip angle of $8^\circ$, $24$ cm field of view. The acquisition matrix is $192 \times 192 \times 166$ in the x, y, and z dimensions and the voxel size is $1.25 \times 1.25 \times 1.2 mm^3$. In-plane, zero-filled reconstruction (i.e., sinc interpolation) generate a $256 \times 256$ matrix for a reconstructed voxel size of $0.9375 \times 0.9375 \times 1.2 mm^3$.

The hippocampal substructures are segmented with FIRST~\citep{patenaude:2011a} based on the T1-weighted MRI scans, and hippocampal surfaces are automatically reconstructed based on the segmentations~\citep{Han:2003TPAMI,Lorensen:MarchingCubes}. To overcome the noise caused by the image scanning and the partial volume effects, a smoothing process consisted of two steps, i.e., progressive meshes~\citep{Hoppe1996ProgressiveM} and loop subdivision\citep{loop1987smooth}, is applied to the reconstructed hippocampal surfaces. A nonlinear surface registration method, inverse consistent fluid registration method~\citep{shi:nimg13}, is applied to register individual hippocampal surfaces to a standard template surface by constrained harmonic maps. After the registration,  surface deformation statistics are computed to obtain the radial distance (RD)~\citep{Pizer:TMI99,Styner:PNAS06,Thompson:nimg04} represented as the shortest distance from each surface vertex to the medical axis of a tube-shape surface. For the following subspace decomposition analysis, we represent an individual's RD features as a column vector and concatenate them to construct group morphometry observation matrices. 

\subsection{Low-Rank and Sparse Subspace Decomposition Algorithm}\label{sec:SPCP}

Based on the robust principal component analysis (RPCA)~\citep{Candes:JACM11}, the low-rank component $L$ and the sparse component $S$ in a morphological observation matrix $A\in\Re^{m\times n}$ ($m$ is the dimension of the biomarker and $n$ is the number of the subjects) can be extracted through a convex optimization procedure. The low-rank component $L$ is the group common structure contributed by all the individuals in the same group, and the sparse component $S$ is the sparse characteristics which belong to specific individual subjects. In practice, the observation matrix is usually corrupted by noise $N$ which can affect every data matrix entry and lead to analysis errors. To solve this problem and achieve a stable separation, a convex model called stable principal component pursuit (SPCP)~\citep{Zhou:ISIT10} has been proposed. This model may guarantee a stable recovery of $L$ and $S$ with an error bound proportional to the overall noise magnitude $\parallel N \parallel_F$. The solution to SPCP can be formulated by the following equation,
\begin{equation}\label{eqn:SPCP}
\min\limits_{L,S} \Vert L\Vert_* + \lambda\Vert S\Vert_1 \quad s.t.:~ \Vert L+S-A\Vert_F \leq \epsilon
\end{equation}

\noindent In Eq.~\ref{eqn:SPCP}, $\Vert \cdot \Vert_*$ is the nuclear norm and $\Vert \cdot \Vert_1$ is the $l_1$-norm, given by $\Vert L\Vert_*=\Sigma_i\sigma_i(L)$, $\Vert S \Vert_1=\Sigma_{i,j}\mid s_{i,j}\mid$ respectively, where $\sigma_i(L)$ is the vector of singular values of $L$, the $\lambda$ controls the relative importance of the low-rank component $L$ vs. the sparse component $S$. The noise $N$ is implicit in the equation, i.e., $L+S+N=A$. Since SPCP algorithm involves minimizing a combination of $l_1$-norm and nuclear norm, the singular value decomposition (SVD) step included in the nuclear norm makes the convergence of SPCP optimization problem very slow, especially for large-scale observation matrices~\citep{Tseng:SIAM08,Lin:2009}. To reduce the SVD computation cost for a large-scale matrix, we adopt a low-rank matrix factorization mechanism~\citep{Liu:Neural12} to factorize a large-sized matrix into the product of two small-sized matrices. Because the group common structure $L$ is of low-rank, $L$ can be factorized into the product of two small-sized matrices, i.e., $L=UV, U\in\Re^{m\times r}, V\in\Re^{r\times n}, r\ll m, r\ll n$. Then we obtain $\Vert L\Vert_*=\Vert UV\Vert_*=\Vert V\Vert_*$ by enforcing $U$ to be column orthonormal, i.e., $U^TU=I$, $I$ is the identity matrix. Thus, Eq.~\ref{eqn:SPCP} can be transformed into an optimization problem constrained on a small-scale matrix,
\begin{equation}\label{eqn:SPCPTran}
\min\limits_{U,V,S} \Vert V\Vert_* + \lambda\Vert S\Vert_1 \quad s.t.:~ \Vert UV+S-A\Vert_F \leq \epsilon, \quad L=UV, \quad U^TU=I.
\end{equation}
Since any low-rank matrix $L$ can be factorized into a product of two small-sized matrices $L=UV$ with $U^TU=I$, this transformation significantly speeds up the SPCP optimization for large-scale matrices.

To extract the sparse component $S$ without loss of the original spatial information, we define $S$ as continuous patches in local regions. We further propose a novel formulation which considers the original spatial distribution of the sparse morphological component $S$. Supposing we have an observation data matrix $A \in \Re^{m \times n}$, and $A=\{a_1,...,a_i,...,a_n \}$ where $a_i$ denotes the column vector of the original morphological measurements of $i$-th subject. In order to detect the local continuous characteristics, we enforce a local sparse measure$\Vert S\Vert_{\Omega,1}$ for every column in the observation matrix, defined as follows:
\begin{equation}\label{eqn:LCC}
\left\{
\begin{array}{ll}
A(p,i) \rightarrow X^i_q, \quad \bar{X}^i_q=(\sum_{\Omega_q}(X^i_{q})^2)^{\frac{1}{2}}, \quad \bar{X}^i_q\rightarrow \bar{A}(p,i) \\
\Vert S\Vert_{\Omega,1}=A(p,i)\odot \mathcal{T}_{\mu}|\bar{A}(p,i)| \\
\end{array}
\right.
\end{equation}

\noindent where $p$ is the number of the row in the observation matrix and $q$ is the vertex sequence number in the triangular mesh of the specific subject. And $\Omega_q$ denotes the one-ring area centered on the vertex $q$, $\odot$ is the Hadamard product, $\mathcal {T}_{\mu}$ is the shrinking function and $\mu$ is the threshold. By using $\Vert S\Vert_{\Omega,1}$ to replace the $l_1$-norm constraint in Eq.~\ref{eqn:SPCPTran}, and representing the local spatial continuous characteristics of the salient features, Eq.~\ref{eqn:SPCPTran} can be modified as,
\begin{equation}\label{eqn:SPCPTranmodi}
\min\limits_{U,V,S} \Vert V\Vert_* + \lambda\Vert S\Vert_{\Omega,1} \quad s.t.:~ \Vert UV+S-A\Vert_F \leq \epsilon, \quad L=UV, \quad U^TU=I
\end{equation}

Given the well-known alternating direction method and the Douglas-Rachford operator splitting (DR) method~\citep{Douglas:mathematical}, the minimization task of the Eq.~\ref{eqn:SPCPTranmodi} can be decomposed into three simpler tasks which solve the variables $L, S, N$ separately in a consecutive order. Let $Y$ be the Lagrange multiplier associated with the linear constraint in Eq.~\ref{eqn:SPCPTranmodi}, the augmented Lagrangian function of Eq.~\ref{eqn:SPCPTranmodi} is:
\begin{equation}\label{eqn:LagSPCP}
L_{\mathcal{A}}(V,S,N):=\Vert V\Vert_*+\lambda\Vert S\Vert_{\Omega,1} -<Y, UV+S+N-A>+ \frac{\beta}{2}\Vert UV+S+N-A\Vert^{2}_F
\end{equation}

Based on the DR method, let $\beta_k >0 $, the according iteration procedure for Eq.~\ref{eqn:LagSPCP} is described as follows:
\begin{equation}\label{eqn:itration}
\left\{
\begin{array}{llll}
N^{k+1}=argmin_{N \in \mathbb{B}} \frac{\beta_k}{2} \Vert N+U^kV^k+S^k-\frac{1 }{\beta_k}Y^k-A\Vert^{2}_F \\
S^{k+1}=argmin_{N \in \Re^{m \times n}} \lambda\Vert S\Vert_{\Omega,1}+ \frac{\beta_k}{2}\Vert S+U^kV^k+N^{k+1}-\frac{1 }{\beta_k}Y^k-A\Vert^{2}_F \\
L^{k+1}=argmin_{V \in \Re^{r\times n}} \Vert V\Vert_*+ \frac{\beta_k}{2}\Vert UV+S^{k+1}+N^{k+1}-\frac{1 }{\beta_k}Y^k-A\Vert^{2}_F \\
Y^{k+1}= Y^{k}-\beta_k (U^{k+1}V^{k+1}+S^{k+1}+N^{k+1}-A) \\
\end{array}
\right.
\end{equation}

The generated subproblems in Eq.~\ref{eqn:itration} admit analytic solutions. The first subproblem involving $N^{k+1}$ can be solved explicitly via:
\begin{equation}\label{eqn:resolveN}
N^{k+1}=\frac{min\{\Vert N^k \Vert_F, \epsilon \}}{\Vert N^k \Vert_F}N^k
\end{equation}

\noindent where $N^{k}= \frac{1 }{\beta_k}Y^k+A-U^kV^k-S^k$. Note that the computation complexity for this subproblem is $O(mn)$. For the second subproblem involving $S^{k+1}$ in Eq.~\ref{eqn:itration}, it is  equivalent to solve the following equation.
\begin{equation}\label{eqn:resolveSpre}
\min \limits_{S} \lambda\Vert S\Vert_{\Omega,1}+\frac{\beta_k}{2}\Vert \frac{1 }{\beta_k}Y^k+A-U^kV^k-N^{k+1}-S\Vert^{2}_F
\end{equation}
Denoting $G_S^k=\frac{1 }{\beta_k}Y^k+A-U^kV^k-N^{k+1}$, and substituting $G_S^k(p,i)$ for $A(p,i)$ in Eq.~\ref{eqn:LCC}. The following formulation gives a closed-form solution of the minimization with respect to
\begin{equation}\label{eqn:resolveS}
S^{k+1}(p,i)= G_S^k(p,i) \max (0,1-\frac{\lambda / \beta_k }{\bar G_S^k(p,i) })
\end{equation}
Here $\bar G_S^k(p,i)$ is the result matrix which is the $\bar A(p,i)$ in Eq.~\ref{eqn:LCC}. Therefore, Eq.~\ref{eqn:resolveS} sets the pixels of $S(p,i)$ to be zero value if the corresponding $\bar G_S^k(p,i)$ is less than $\lambda / \beta_k$, or shrinks the corresponding pixel by a factor $1-\frac{\lambda / \beta_k}{\bar G_S^k(p,i)}$. After an iteration of $S$ update, we obtain $S^{k+1},\quad N^{k+1}$ terms. Denote $G_L^k=\frac{1 }{\beta_k}Y^k+A-E^{k+1}-N^{k+1}$, and the known condition $(U^k)^TU^k=I, L^k=U^kV^k$. To acquire the $U^{k+1}$ with column orthonormal and $V^{k+1}$ with a small scale, we chose two-steps. First, we fix the $V^k$ and compute the $U^{k+1}$. According to~\cite{Higham:1995}, the update of $U^k$ is given by the following equation,
\begin{equation}\label{eqn:resolveU}
U^{k+1}= \mathcal{P}(G_L^k (V^k)^T)
\end{equation}

For a given matrix $Z=G_L^k(V^k)^T$, SVD can be applied to $Z$ and the result is $Z=P \Sigma Q$. Therefore the operator $\mathcal{P}$ can be defined as $\mathcal{P}(Z)=PQ^T$. It can be verified that $\mathcal{P}(Z)$ is always column orthonormal (i.e., $(\mathcal{P}(Z))^T \mathcal{P}(Z)=I$ ). Through Eq.~\ref{eqn:resolveU}, we can get the $U^{k+1}$ by satisfying the condition (i.e., $\Vert L^{k+1}\Vert_*=\Vert U^{k+1}V^{k+1}\Vert_*=\Vert V^{k+1}\Vert_*$ ). Second, fixing $U^{k+1}$ and computing the $V^{k+1}$, the update of $V_k$ is as follows,
\begin{equation}\label{eqn:resolveV}
V^{k+1}= \mathcal{Q}_{\mu}((U^{k+1})^T G^k_L)
\end{equation}

For a given matrix $T=(U^{k+1})^T G^k_L$, $\mu>0$, the solution of minimizing of the nuclear norm of $T$ is given by
$\mathcal{Q}_{\mu}(T) $, defined by
\begin{equation}\label{eqn:resolveV1}
\mathcal{Q}_{\mu}(T):=J diag(S_{\mu}(\Sigma))K
\end{equation}
where $J $, $K $ and $\Sigma $ are obtained by the singular value decomposition (SVD) of $T$, and $S_{\mu}(\Sigma):=sign(\Sigma_{i,i})\cdot max(0,abs(\Sigma_{i,i})-\mu)$~\citep{Cai:Siam2008}. Here $\mu=1/ \beta_{k+1}$.
Thus, according to Eq.~\ref{eqn:resolveU} and Eq.~\ref{eqn:resolveV} , the solution of $L_{k+1}$ in Eq.~\ref{eqn:itration} can also be solved by
\begin{equation}\label{eqn:resolveUV}
L^{k+1}=U^{k+1}V^{k+1}
\end{equation}

With Eq.~\ref{eqn:resolveU} and Eq.~\ref{eqn:resolveV}, the SVD computation complexity is reduced from $O(mn\min (mn))$ to $O(mnr)$. In addition, we set the parameters in Eq.~\ref{eqn:LagSPCP} as $\lambda=\frac{1}{\sqrt{\max \{m,n\}}}$, where $m$ and $n$ represent the numbers of rows and columns in the observation data matrix. The parameter $\beta_0=\frac{1.25}{\sigma_L}, \beta_{k+1}=\alpha \cdot \beta_{k}$. The $\sigma_L$ is the largest singular value of the observation data matirx and $\alpha \in (1,2)$. And the parameter $\epsilon$ in Eq.~\ref{eqn:SPCP} is set as $\sqrt{(\min (m,n)+\sqrt{8} \cdot \min (m,n))}\cdot \delta$~\citep{Tao:SIAM11}, $\delta$ is the standard deviation of the observation data matrix. Finally, the algorithm is terminated when
$\frac {\Vert (U_{k+1}V_{k+1},S_{k+1})-(U_{k}V_{k},S_{k}) \Vert_F}{\Vert (U_{k}V_{k},S_{k})\Vert_F+1} \leq \tau \cdot \delta $. Based on the above description, we can use the proposed method to efficiently and accurately recover the low-rank component and the sparse component with a local continuous constraint from the large-scale data matrix in a complicated environment. In particular, we can use the low-rank component, i.e., where $L$ denotes the group common structure, to study the group differences without considering the individual specific diversities in $S$.

\subsection{Univariate Morphometry Index (UMI)}\label{sec:MSSI}

Taking the morphological RD measures on the hippocampal surfaces as features, we apply the subspace decomposition algorithm to analyze  data from both $A\beta+$ AD and $A\beta-$ CU groups and  obtain the essential group common morphological structures from the individuals of the same group, i.e., $L$s of $A\beta+$ AD group and $A\beta-$ CU group, while discarding the within-group morphological differences and the noise. Using the two group common morphological structures, the ROIs are generated through group difference analysis. This means that the generated ROIs reflect the hippocampal morphological structure changes from CU to AD. Next, we obtain the voxel-wise AD atrophy degrees by normalizing the mean differences between the $L$ of A$\beta+$ AD group and the $L$ of A$\beta-$ CU group in the predefined ROIs. The individual atrophy degree can be computed in the predefined ROIs by normalizing the differences between the RD vector of an individual subject and the low-rank component mean of the A$\beta-$ CU group. Last, the UMIs can be obtained through the voxel-wise summation across all the voxels on the predefined ROIs with the production of AD atrophy degrees and the individual atrophy degree:
\begin{equation}\label{eqn:MSSI}
\textrm{UMI}=\frac {\sum_{i=1}^r (D_{T_i} \cdot D_{W_i})}{100}
\end{equation}

\noindent where $r$ is the total number of voxels in the predefined ROI, $D_{T_i}$ denotes the atrophy degree at voxel $i$ of the testing individual $T$ (defined as the mean of the A$\beta$ negative CU group subtracted by the individual RD on the predefined ROI), and $D_{W_i}$ is the atrophy degree at voxel $i$ (defined as the difference of the mean of A$\beta+$ AD group and the mean of A$\beta-$ CU group on the predefined ROI). From Eq.~\ref{eqn:MSSI}, we see that the UMIs measure how similar the individual atrophy degree is when compared to the AD atrophy degree in the selected ROIs, similar to our previous work~\citep{Chen:HCI11}. In other words, the greater the UMIs, the closer the individual atrophy degree from the AD morphological characteristics.

\subsection{Assessments of Univariate Morphometry Index}\label{sec:assessment}

\subsubsection{Group Comparisons}
We compare the group mean differences for the UMI results of different longitudinal A$\beta+$ groups based on the paired $t$-test, including longitudinal A$\beta+$ AD, A$\beta+$ MCI and A$\beta+$ CU groups. All subjects of each longitudinal group underwent two tests, including the baseline test and a 24-months test. We use Cohen's d test to evaluate the effect size. Cohen's d test takes the difference in means between two groups and is devided by the pooled standard deviation of the groups. For MCI converters and non-converters within 18-months period, we analyze the group differences of the UMI measures, Mini-Mental State Examination (MMSE)~\citep{folstein:1975} scores, Auditory Verbal Learning Test Total (AVLT-Total)~\citep{AVLT} scores and hippocampal volume measurements by using analysis of variance (ANOVA) method to compare their statistical discrimination abilities .

\subsubsection{Minimum Sample Size Estimation for Longitudinal Study}
To assess the statistical power of the defined UMI, we study the minimum sample size estimation, i.e., with 80$\%$ power and a 25$\%$ reduction in the mean annual change, using a two-sided test and standard significance level $\alpha-0.05$ for a hypothetical two-arm study (treatment vs. placebo, as advocated by the ADNI Biostatistics Core)~\citep{Beckett:statcore} for longitudinal (baseline and 24-months follow-up) studies. The minimum sample size is computed from the below formula:
\begin{equation}\label{eqn:n80}
\textrm{minimum sample size}=C \frac{\sigma^2}{(m_1-m_2)^2}
\end{equation}
\noindent where $\sigma$ denotes the standard deviation of the biomarkers' changes, $m_1$ and $m_2$ refer to the mean value of the biomarkers at the time $t_1$ and $t_2$ of the longitudinal data. $C$ is a constant. If the minimum sample size derived from a biomarker $B_1$ was smaller than that of another biomarker $B_2$, we may conclude that biomarker $B_1$ has more differentiation ability than biomarker $B_2$. It suggests that minimum sample size is an indicator for characterizing the statistical power of biomarkers.

\subsubsection{Prediction Rates of Progression from MCI to probable AD}
We compare UMIs, hippocampal volume measurements and AVLT-Total scores in their abilities to predict rates of conversion from MCI to AD within the 18-months period after baseline. We initially use Receiver Operating Characteristic (ROC) analyses to determine the measurement cut-offs with optimal sensitivity and specificity for distinguishing between the MCI converters and those who are stable. The nearest point method is used to find the cut-offs on the ROC curve closest to (0,1), that is, the point with optimal sensitivity and specificity. The cut-offs are used to compare the MCI patients with or without a positive test score (i.e., those with or without a higher UMI, lower hippocampal volume or lower AVLT-Total score) in the hazard ratio of converting from MCI to AD within 18-months period. We also report area under the curve (AUC), which is an effective and combined measure of sensitivity and specificity that describe the inherent validity of a classification test. Thus, for each biomarker of interest, univariate Cox proportional hazard model is constructed to determine the hazard ratio for progression to AD within the 18-months period in MCI patients who are classified into subjects with scores above the cut-off value (positive biomarker) in comparison with those with scores below (negative biomarker). And Kaplan-Meier plots are used to describe the ability of each biomarker of interest to distinguish between those MCI patients with and without a positive biomarker.


\section{Experimental Results}\label{sec:3}

\subsection{Computing Univariate Morphometric Index (UMI) with Hippocampal Surface Features}\label{sec:ROIextract}

\begin{figure}[htp]
\begin{center}
\includegraphics[width=0.8\textwidth]{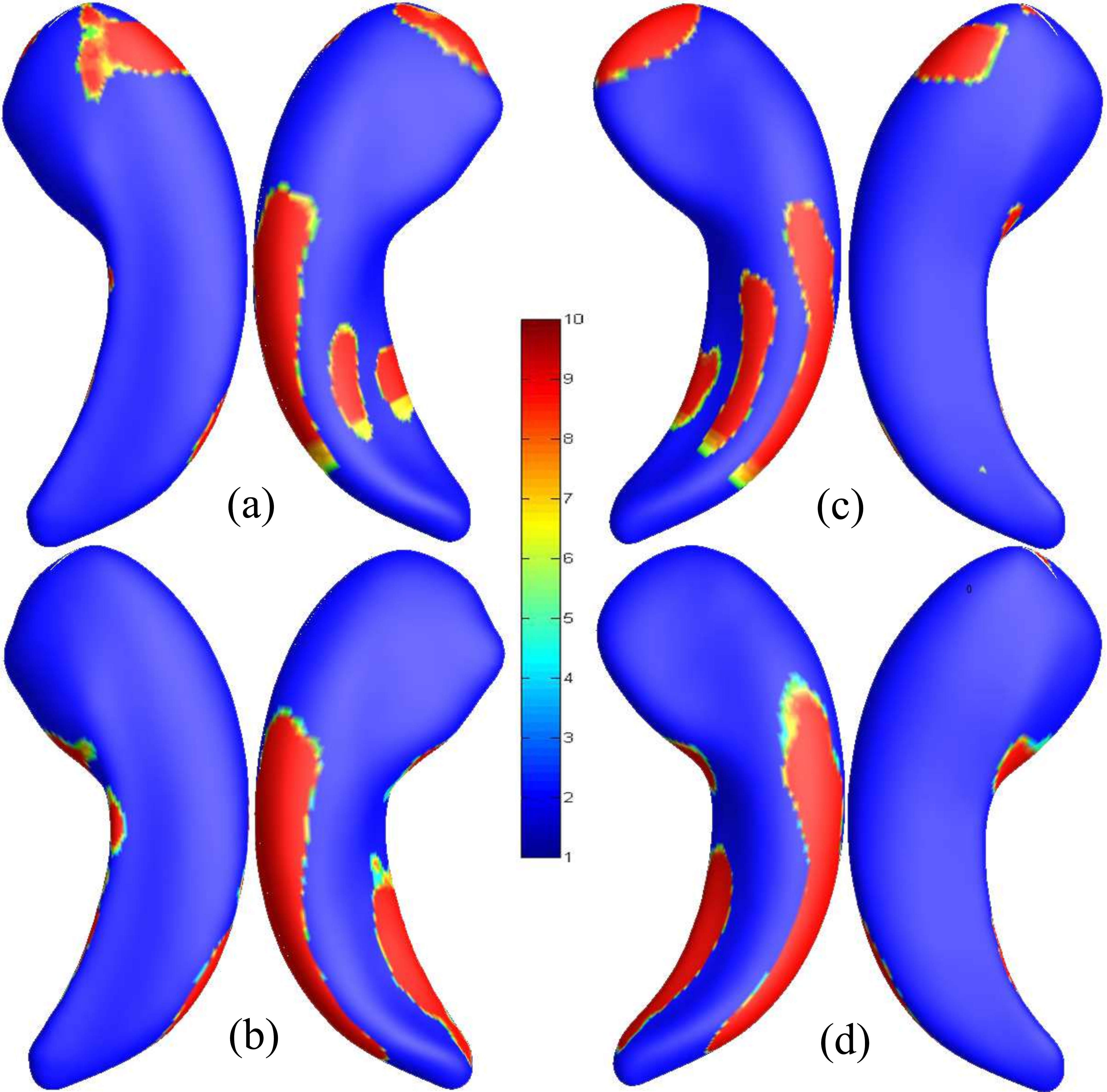}
\caption{The extracted ROI results for the $A\beta+$ AD and $A\beta-$ CU group differences based on the low-rank data and the radial distance (RD) data. (a) and (b) are the statistical ROI results for the left hippocampus (LH). (c) and (d) are the statistical ROI results for the right hippocampus (RH). And (a) and (c) are the statistical ROI results based on the low-rank components. (b) and (d) are the statistical ROI based on the raw RD data. All statistical ROI results are obtained by the permutation $t$-test (p-value$<0.00001$). Non-blue colors indicate the number of times a vertex is selected.}
\label{fig:ROI}
\end{center}
\end{figure}

To extract the significant morphological changes induced by AD, we apply the subspace decomposition to a cohort consisting of patients with dementia due to AD (defined $A\beta+$ AD) and A$\beta-$ CU subjects. In the ADNI dataset, there are $422$ subjects including $151$ A$\beta+$ AD patients and $271$ A$\beta-$ CU subjects (Table~\ref{tab:demographic}). Demographic and clinical data are compared using a one-way analysis of variance, and the gender data are analyzed by a $\chi^2$ test ~\citep{Crivello:Bio2014}. Table~\ref{tab:demographic} indicates that the factors of age and gender of the two groups are matched, while the MMSE is significantly different between these two groups.
\begin{table}[t]
\begin{center}
\begin{tabular}{cccc}
\hline\hline
  & $A\beta+$ AD & $A\beta-$ CU &  Inferential Statistics\\
\hline\hline
Sample Size & $151$ & $271$ &   \\
Gender(M/F) & $79/72$ & $132/139$  & $\chi^2=0.51; p=0.48 $ \\
Age & $74.58\pm7.88 $ & $74.02\pm6.68$ &  $F=0.60; p=0.44$ \\
MMSE & $22.60\pm3.09$ & $28.94\pm2.18$ &  $F=602.03; p<0.0001 $ \\
\hline
\end{tabular}
\caption{Demographic information of A$\beta+$ AD subjects and A$\beta-$ CU subjects}\label{tab:demographic}
\end{center}
\end{table}

 When the individual RD features belonging to the same group are stacked into columns to generate a morphological observation matrix, the observation matrix is low-rank, and the low-rank components correspond to the group common morphological structure. We use the generated common morphological structures of A$\beta+$ AD group and A$\beta-$ CU group to form ROIs which represent the regions significantly affected by ADD. To validate if generated ROIs are consistent each time, we randomly divide the subjects of each group into 10 folds and chose 9 folds from each group as the training sets. This process is repeated 10 times. We apply our subspace decomposition algorithm to extract the low-rank component $L$ of $A\beta+$ AD and $A\beta-$ CU groups. The total performance indicators for left hippocampus (LH) and right hippocampus (RH) of the two groups based on our algorithm are shown in Table ~\ref{tab:performance}. Here we set the parameter $\lambda$ as 0.0082.

\begin{table}[t]
\begin{center}
\begin{tabular}{ccccccc}
\hline\hline
 &   & $\beta_0$ & $\delta$ 	&  $rank(L)$  & $\Vert S\Vert_{\Omega,1}$   & $\Vert L+S-A \Vert / \Vert A\Vert$ \\
\hline\hline
\multirow{2}*{LH}	   & A$\beta$- CU &$0.08$ &$0.06\sim0.08$ & $11\sim12$ & $267\sim280$  & $0.080$\\
    & A$\beta$+ AD &$0.11$ &$0.05\sim0.07$ & $9\sim10$ &  $365\sim388$ & $0.084$\\

\hline
\multirow{2}*{RH}     & A$\beta$- CU &$0.08$ &$0.06\sim0.09$ & $11\sim12$ & $285\sim301$  & $0.081$\\
    & A$\beta$+ AD &$0.11$ &$0.05\sim0.07$ & $9\sim10$ &  $411\sim425$ & $0.089$\\

\hline\\

\end{tabular}
\caption{Performance indicators of the subspace decomposition algorithm for the A$\beta+$ AD group and the A$\beta-$ CU groups. LH and RH indicate left hippocampus and right hippocampus, respectively.}\label{tab:performance}
\end{center}
\end{table}

As shown in Table~\ref{tab:performance}, the computed $\beta_0$, $\delta$, $rank(V)$ and $\Vert L+S-D \Vert / \Vert D\Vert$ parameters of the same group are similar whether on the LH or the RH. The rank of $L$ is obtained by counting the number of the eigenvalues of $L$ whose values are greater than three times the $\delta$. After extracting the $L$ and $S$ from the training data each time, we define the ROIs by permutation tests~\citep{Gill:permutation07} between two groups based on the low-rank component $L$. The permutation $t$-test is a type of statistical significance test. The distribution of the test statistics under the null hypothesis is obtained by calculating all possible values under rearrangements of the labels on the observed data points. As the training samples are taken randomly each time (the changing rate is about $10\%$), each obtained ROI will be inconsistent if the extracted $L$ components are not robust. We compare the extracted ROIs based on the $L$ components with the RD data. The generated ROIs are the vertices whose permutation $t$-test $p$-values represent the group differences smaller than $0.00001$. The results of the generated ROIs are shown in Fig.~\ref{fig:ROI}, where (a) and (b) are the statistical ROI results for the LH, (c) and (d) are the statistical ROI results for the RH. Fig.~\ref{fig:ROI} (a) and (c) are the statistical ROI results based on the low-rank components. (b) and (d) are the statistical ROIs based on the raw RD data. Non-blue colors indicate the number of times a vertex is selected. The red color denotes that the vertex is selected for 10 times after 10 ROI generation processes. From the results, we can see that the selected ROIs based on the low-rank components are more robust than the ones based on the raw RD data. For the left hippocampus, the vertices selected for 10 times account for 73$\%$ of the total number of ROI vertices based on the low-rank components, while it is 51$\%$ based on the raw RD data. It suggests that the low-rank components may represent the essential common structure from the subjects without the interference of the unique individual structure and the noise.

All the experiments are performed with an Intel Core i7 personal computer, with 3.40 GHz CPU, 16GB RAM and MATLAB 2016 installed in the Win 7 operating system. The execution time for extracting low-rank components of right hippocampus (RH) and left hippocampus (LH) from $151$ $A\beta+$ patients and $271$ $A\beta-$ subjects is approximately $170.24$ seconds. The execution time for generating ROIs of RH and LH by permutation $t$-tests between these two groups is approximately $7441.64$ seconds when the number of random permutation processes is set as $5,000$ based on the Monte Carlo method. Based on the obtained low-rank components and ROIs, the execution time to compute UMI of a new subject by using Eq.~\ref{eqn:MSSI} is approximately $0.12$ seconds on average. Since the ROI generation is usually done once during the training state, our relatively short UMI computation time on a new subject ($0.12$ seconds) demonstrates the feasibility of our algorithm for online UMI computation.

In order to verify whether the generated UMI results based on the low-rank components have reasonable statistical power for distinguishing different clinical groups, we introduce the UMI based on raw RD measures (UMI-RD) and the volume measures to compare the discrimination powers with the UMIs. The UMIs of the longitudinal A$\beta+$ subjects are computed according to the Eq.~\ref{eqn:MSSI} based on the obtained ROIs shown in Fig.~\ref{fig:ROI} (a) and (c). The UMIs-RDs of the longitudinal A$\beta+$ subjects are computed according to Eq.~\ref{eqn:MSSI} based on the obtained ROIs shown in Fig.~\ref{fig:ROI} (b) and (d). For the clarity of the content in the following sections, if not stated, UMIs refer to those generated based on subspace decomposition and UMI-RDs refer to those generated based on raw RD measures. The volumetric MRI measurements of the hippocampus are generally accepted as one of the best established biomarkers for clinical AD progression research~\citep{Jack:Neuro99}. Similar to prior approaches that used hippocampal volume for AD diagnosis,~\citep[e.g.][]{Pennanen:NIMG04,Chupin:NIMG07,Chupin:HP09}, the hippocampal volume is computed on our smoothed surfaces after they are linearly registered to the MNI imaging space~\citep{patenaude:2011a,shi:nimg13,Dong:NICL19}.

\subsection{Longitudinal Data Analysis}\label{sec:SPLD}

Based on the predefined ROIs and the AD atrophy degrees, we study the statistical differentiation ability of the low-rank components for the longitudinal subjects. In this study, we use $165$ longitudinal $A\beta+$  subjects, including $46$ AD, $55$ MCI and $64$ CU subjects. The demographic characteristic statistics information for the study samples is shown in Table ~\ref{tab:demographic1}. All subjects undergo two tests, including the baseline test and a 24-months test.

\begin{table}[t]
\begin{center}
\begin{tabular}{ccccc}
\hline\hline
& Gender(M/F) 	& Education   & Age    & MMSE\\
\hline\hline
AD baseline& \multirow{2}*{$21/25$}	& \multirow{2}*{$14.33\pm2.13$} & $75.09\pm8.60$  & $22.24\pm1.80$\\
AD m24& 	&  & $77.18\pm8.65$  & $17.88\pm4.46$\\
\hline
MCI baseline& \multirow{2}*{$29/26$}	& \multirow{2}*{$15.73\pm2.31$} & $72.49\pm6.67$  & $27.51\pm1.88$\\
MCI m24& 	&    & $74.52\pm6.46$  & $25.50\pm2.76$\\
\hline
CU baseline& \multirow{2}*{$28/36 $} & \multirow{2}*{$15.87\pm2.25$} & $75.27\pm5.33$  & $29.09\pm0.92$\\
CU m24&  &  & $77.96\pm5.68$  & $28.06\pm1.21$\\
\hline
\end{tabular}
\caption{Demographic information of A$\beta+$ subjects in three clinical groups.}\label{tab:demographic1}
\end{center}
\end{table}

\subsubsection{Group Difference Study}

The obtained UMI, UMI-RD and volume results show that the mean values of the UMIs and UMI-RDs have a upward trend while the mean value of the volume measures have a downward trend from baseline to the 24-months follow-up, among CU to MCI to AD stage. The statistical comparison results of different longitudinal groups based on the UMI, UMI-RD and volume measures are shown in Table~\ref{tab:MSSIresult}. The $p$-value results of different longitudinal groups are computed by two-sided paired $t$-tests. The effect sizes of different longitudinal groups are computed by paired Cohen's d measure.

\begin{table}[t]
\begin{center}
\begin{tabular}{cccc}
\hline\hline
\textbf{Longitudinal Group (bl vs. m24)}  & \textbf{AD}  & \textbf{MCI} & \textbf{CU} \\
\hline\hline

\textbf{ $p$-value of UMI} & \textbf{$ 7.74e-15$ } & $ 3.54e-10$ & $ 1.28e-09$ \\
\textbf{ Effect Size of UMI} & \textbf{$1.39$ } & $ 0.92$ & $ 0.85$ \\ 
\hline\

\textbf{ $p$-value of UMI-RD} & $ 1.52e-03$ & $ 3.70e-03$ & $ 1.29e-02$ \\
\textbf{ Effect Size of UMI-RD} & $0.79$  & $ 0.65$ & $ 0.56$ \\ 
\hline\

\textbf{ $p$-value of Volume} & $ 1.48e-13$  & $ 9.72e-09$ & $ 1.19e-08$ \\
\textbf{ Effect Size of Volume} & $1.31$ & $ 0.90$ & $ 0.74$ \\ 
\hline\hline\
\end{tabular}
\caption{The statistical comparison results, including group difference $p$-value and effect size, of different longitudinal groups based on the UMI, UMI-RD and volume measures.}\label{tab:MSSIresult}
\end{center}

\end{table}

For the longitudinal $A\beta+$ AD, $A\beta+$ MCI and $A\beta+$ CU , the $p$-values and effect sizes for UMIs mean differences are $7.74e-15$ and $1.39$, $3.54e-10$ and $0.92$, $1.28e-09$ and $0.85$, respectively. For UMI-RDs, the $p$-values and effect sizes of the longitudinal $A\beta+$ AD, $A\beta+$ MCI and $A\beta+$ CU are $1.52e-03$ and $0.79$, $3.70e-03$ and $0.65$, $1.29e-02$ and $0.56$, respectively. For hippocampal volume measures, the $p$-values and effect sizes of the longitudinal $A\beta+$ AD, $A\beta+$ MCI and $A\beta+$ CU are $1.48e-13$ and $1.31$, $9.72e-09$ and $0.90$, $1.19e-08$ and $0.74$, respectively. The results indicate that the UMIs may have a stronger discrimination ability to distinguish the longitudinal groups than both the UMI-RDs and the hippocampal volume measures.

\begin{figure}[H]
\begin{center}
\includegraphics[width=0.9\textwidth]{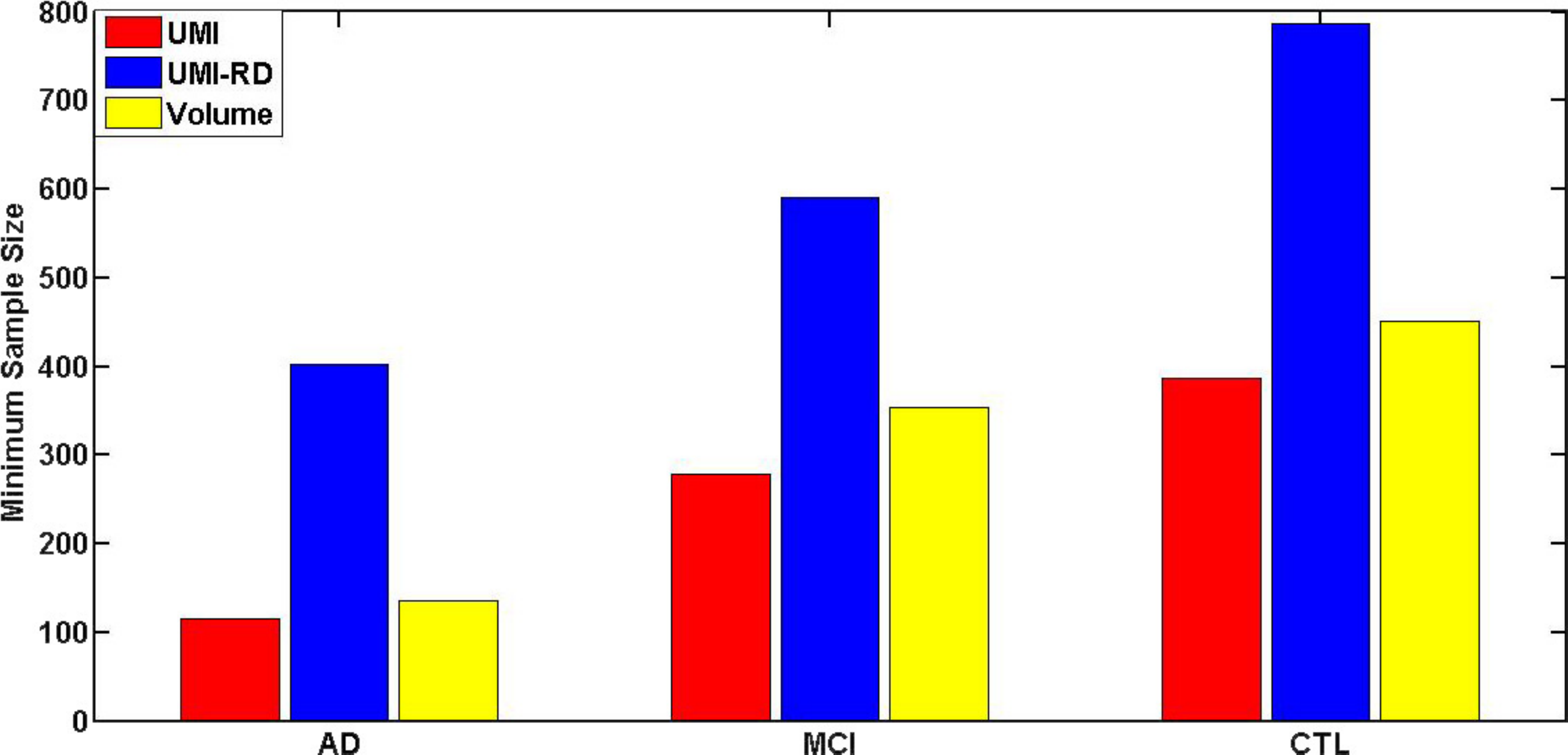}

\caption{The minimum sample size comparisons between UMIs, UMI-RDs and volume measures for AD, MCI and CU longitudinal A$\beta$ groups.}
\label{fig:n80ROI1}
\end{center}
\end{figure}

\subsubsection{Minimum Sample Size Estimation}
Next we will demonstrate the reduction of minimum sample size by using UMI as a biomarker in clinical studies. With Eq.~\ref{eqn:n80}, we estimate the minimum sample sizes of UMIs, UMI-RDs and volume measures. As shown in Fig.~\ref{fig:n80ROI1}, the minimum sample sizes of the UMIs of the longitudinal $A\beta+$ AD, $A\beta+$ MCI and $A\beta+$ CU groups are 116, 279 and 387, respectively. For UMI-RDs, the minimum sample sizes of the longitudinal $A\beta+$ AD, $A\beta+$ MCI and $A\beta+$ CU groups are 402, 590 and 787, respectively. For volume measures, the minimum sample sizes of the longitudinal $A\beta+$ AD, $A\beta+$ MCI and $A\beta+$ CU groups are 136, 352 and 451, respectively. Regardless of whether it is based on the UMIs or the volume measures, the minimum sample sizes of longitudinal $A\beta+$ AD group is smallest, followed by the longitudinal $A\beta+$ MCI group, and the minimum sample sizes of longitudinal $A\beta+$ CU group is largest. It indicates that the morphological changes in the $A\beta+$ AD group are relatively largest, followed by $A\beta+$ MCI group, and the morphological changes in $A\beta+$ CU group are relatively smallest between the baseline and the 24-months follow-up. Meanwhile, the results show that the minimum sample sizes from the volume measures and UMI-RDs are all larger than our UMIs, indicating that the subspace decomposition-based UMIs may have detected the essential morphological changes induced by ADD better than the volume measures and UMI-RD measures.

\subsection{Predicting Clinical Decline in MCI Patients}\label{sec:PRP}
In our experiments, we use 155 MCI patients (Table~\ref{tab:demographicMCI}) containing 59 MCI  subjects who have converted to probable AD and 96 MCI non-converters within 18-months period. Based on the two kinds of predefined ROIs from the low-rank components and raw RD measures of the $A\beta+$ AD and $A\beta-$ CU groups in Section~\ref{sec:ROIextract}, we compute the UMIs and UMI-RDs of MCI converters and MCI non-converters, respectively. The MCI converters are distinguished from the non-converters by the UMIs ($p$-value$<0.0001$), the UMI-RDs ($p$-value$=0.076$), the hippocampal volume measures ($p$-value=0.0015) and the AVLT-Total scores ($p$-value=0.0002). We choose the UMIs, UMI-RDs, hippocampal volume measures and AVLT-Total scores to predict the conversion rates from MCI to AD within 18 months after baseline, respectively.

\begin{figure}[htp]
\begin{center}
\includegraphics[width=0.95\textwidth]{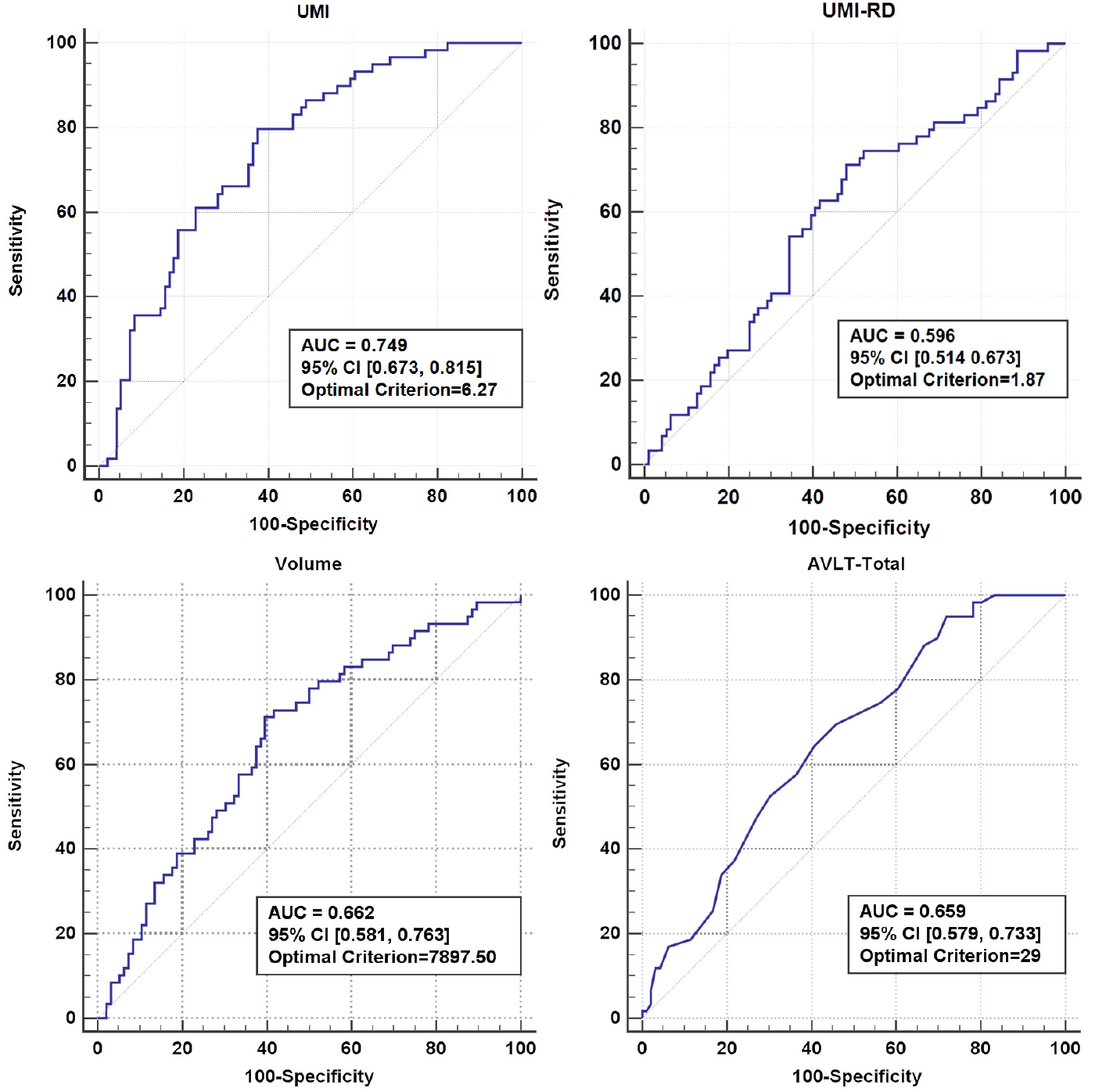}
\caption{The ROC analysis results of UMIs, UMI-RDs, volume measures and AVLT-Total scores.}
\label{fig:cutoff}
\end{center}
\end{figure}

\begin{table}[t]
\begin{center}
\begin{tabular}{cccc}
\hline\hline
  & Converter  & Non-converter & Inferential Statistics\\
\hline\hline
Sample Size & $59$ & $96$ &  \\
Gender(M/F) & $35/24$ & $64/32$ & $\chi^2=0.85; p=0.36  $ \\
Age & $75.46\pm6.65$ & $75.28\pm7.29 $ & $F=0.03; p=0.87$ \\
MMSE & $26.79\pm1.81$ & $27.29\pm1.62$ & $F=3.17; p=0.078 $ \\

\hline
\end{tabular}

\caption{Demographic information of MCI converters and MCI non-converters}\label{tab:demographicMCI}
\end{center}
\end{table}
Then we determine the measurement cut-offs with optimal sensitivity and specificity for distinguishing between the MCI converters and those who are stable based on Receiver Operating Characteristic (ROC) analysis. As shown in Fig.~\ref{fig:cutoff}. The AUC, $95\%$ confidence interval (CI) of AUC and optimal criterion are $0.749$, $[0.673, 0.815]$ and $6.27$ for the UMIs, $0.596$, $[0.514, 0.673]$ and $1.87$ for the UMI-RDs, $0.659$, $0.662$, $[0.581, 0.763]$ and $7897.50$ for volume measures and $[0.579, 0.733]$ and $29$ for AVLT-Total scores, respectively.

\begin{table}[t]
\begin{center}
\begin{tabular}{cccc}
\hline\hline
& $\beta$  & $HR$ ($95\%$ CI) & $p$-value\\
\hline\hline
UMI & $1.5$ & $4.3 \quad (2.3-8.2)$ & $5.8e-06 $ \\
UMI-RD & $0.73$ & $2.1 \quad (1.2-3.7)$ & $1.1e-02 $ \\
Volume & $0.99$ & $2.7 \quad (1.5-4.7)$ & $4.5e-04 $ \\
AVLT-Total & $0.69$ & $2 \quad (1.2-3.3)$ & $9.1e-03 $ \\
\hline
\end{tabular}
\caption{Estimated hazards ratios ($HRs$) of 18-months conversion to AD in MCI patients with positive versus negative biomarkers using univariate Cox analyses.}\label{tab:HRestimate}
\end{center}
\end{table}

Based on the optimal criteria of the UMIs, UMI-RDs, volume measures and AVLT-Total scores, four additional labels are added to the MCI-converter and MCI-nonconverter subjects, i.e., High/Low UMI (HU/LU), High/Low UMI-RD (HU-RD/LU-RD), High/Low volume (HV/LV) and High/Low AVLT-Total (HA/LA). In the view of the effects of AD, it may increase the UMI and UMI-RD and decrease the hippocampal volume and AVLT-Total scores. We call HU, HU-RD, LV and LA as positive biomarkers, and LU, LU-RD, HV and HA as negative biomarkers. After separately fitting the Cox regression model~\citep{Coxmodel} using the four labels above, we compute hazard ratios of 18-months conversion to AD in MCI patients with positive biomarkers versus negative biomarkers. The results in Table~\ref{tab:HRestimate} show the regression beta coefficients ($\beta$), the hazard ratios (HRs) and statistical significance ($p$-value) for UMIs, UMI-RDs, volume measures and AVLT-Total scores within the 18-months observation period. The UMIs, UMI-RDs, hippocampal volume measures, and AVLT-Total scores remain significant for predicting conversion from MCI to AD. In particular, when the MCI patients with LU, LU-RD, HA and HV negative biomarkers are chosen as the reference groups, the MCI patients with HU positive biomarkers had the highest HRs of conversion to AD ($4.3$ $95\%$ CI=$2.3-8.2$), followed by LV ($2.7$ $95\%$ CI=$1.5-4.7$), HU-RD ($2.1$ $95\%$ CI=$1.2-3.7$) and LA ($2$ $95\%$ CI=$1.2-3.3$).

\begin{figure}[htp]
\begin{center}
\includegraphics[width=0.6\textwidth]{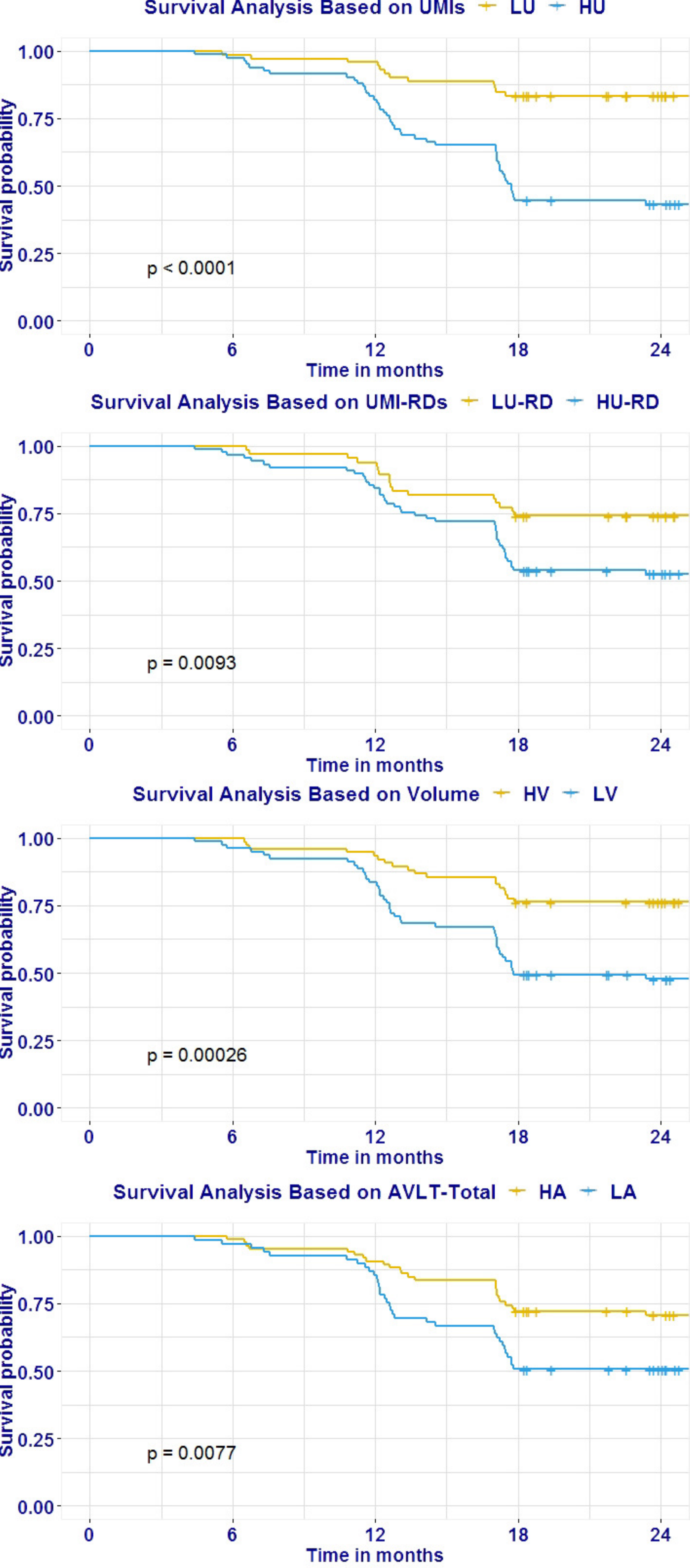}
\caption{The survival probability analysis for progression to AD in MCI patients based on UMI, UMI-RD, hippocampal volume and AVLT-Total measures.}
\label{fig:survival}
\end{center}
\end{figure}

In addition, we compute survival probabilities for progression to AD in MCI patients who are classified into positive or negative biomarkers by fitting the Kaplan-Meier curves. The survival probablities of the MCI patients based on UMIs, UMI-RDs, hippocampus volume measures and AVLT-Total scores are shown in Fig.~\ref{fig:survival}. In each plot, the two lines represent survival curves of the two groups with positive and negative biomarkers, i.e., HU vs. LU, HU-RD vs. LU-RD, HV vs. Lv and HA vs. LA. The vertical lines in each plot indicate the censored data, their corresponding x values indicate the time at which censoring occurred. Here the log-rank test is used to compare the survival group differences based on $\chi^2$ test. A result with $p$-value $<$ 0.05 indicates that the two groups are significantly different in terms of survival time. From the results in Fig.~\ref{fig:survival}, we can conclude that the two group survival differences based on the four labels are significant, that is, the patients with LU, LU-RD, HV and HA have a significantly better prognosis compared to patients with HU, HU-RD, LV and LA. Moreover, the largest survival group difference is found in UMIs, followed by hippocampal volume measures UMI-RDs and AVLT-Total scores. For example, at the time point of 18-months, the probability of survival is approximately 0.45, $95\%$ CI [0.35, 0.57] for HU group and 0.83, $95\%$ CI [0.75, 0.92] for LU group, 0.54, $95\%$ CI [0.45, 0.65] for HU-RD group and 0.74, $95\%$ CI [0.64, 0.86] for LU-RD group, 0.49, $95\%$ CI [0.40, 0.62] for LV group and 0.76, $95\%$ CI [0.67, 0.87] for HV group, 0.52, $95\%$ CI [0.41, 0.65] for LA group and 0.72, $95\%$ CI [0.63, 0.82] for HA group. This may also indicate that the UMIs based on subspace decomposition have stronger discrimination power for survival analysis.

\section{Discussion}
Our current work has two main findings. First, it is possible to develop and apply a low-rank and sparse subspace decomposition approach to solve UNB problems. To develop a robust and effective UNB, an important question is how to accurately extract group common structures with the consideration of recovering sparse components within a feature space corrupted by strong group variances and noise. In this work, we propose to take a subspace decomposition approach and improve the RPCA method~\citep{Candes:JACM11} by adopting a low-rank matrix factorization mechanism~\citep{Liu:Neural12}, imposing regularization constraints into the sparse component to encode spatial connectivity in the original 3D morphometry features and proposing an efficient numerical scheme to solve the formulated optimization problem. We identify the ROIs by studying the group differences between common structures of pathology-confirmed AD group ($A\beta+$ AD) and $A\beta-$ CU groups. Based on the ROIs, we further define our UMI by summarizing an individual morphological structure information in these ROIs. The resulted UMI may sensitively identify abnormal degrees induced by neurodegenerative diseases without the interference of individual morphological discrepancies and noise. Second, although low-rankness and sparsity techniques~\citep{Hastie:Sparsity15,Vidal:Book16,Lin:lowrank17} are widely adopted in computer vision and medical imaging research~\citep[e.g.][]{Wang:BioInf12,Wang:JCB18}, to our knowledge, it is the first work that apply them to surface morphometry features for UNB generations. Our current experimental results in longitudinal data analysis and survival analysis show that surface hippocampal morphometry features, i.e., radial distance~\citep{Pizer:TMI99,Thompson:nimg04,Styner:PNAS06}, lead to superior results than the traditional hippocampal volume measures~\citep{Jack:Neuro99,Cortechslab}.
Similar to previous work~\citep{Thompson:nimg04,Apostolova:NA10,Qiu:TIP10,Costafreda:NIMG11,Dong:NICL19}, it also verifies that surface-based brain subcortical analysis may be useful to capture AD-induced morphological changes.

\subsection{Influence of Subspace Decomposition Parameters }\label{sec:ISDP}

\begin{figure}[t]
\begin{center}
\includegraphics[width=1\textwidth]{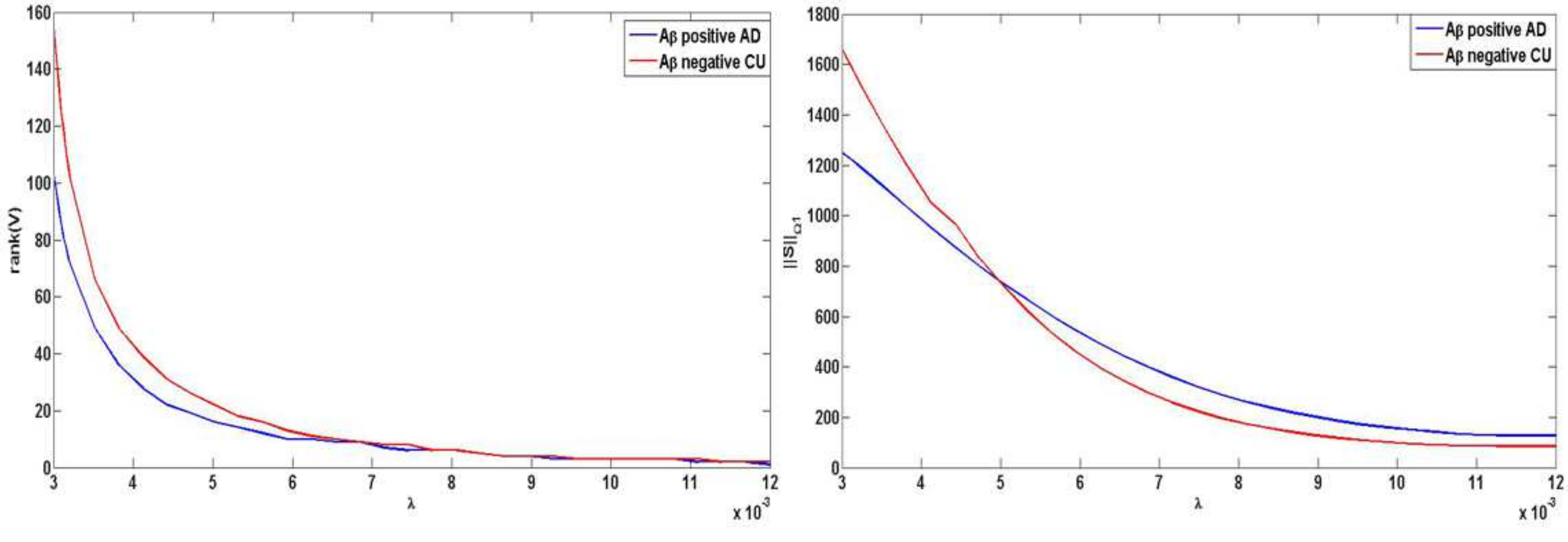}
\caption{The $rank(V)$ and $\Vert S \Vert_{\Omega,1}$ comparisons between $A\beta+$ AD and $A\beta-$ CU groups with the different $\lambda$}
\label{fig:difflam}
\end{center}
\end{figure}

In Eq.~\ref{eqn:SPCPTranmodi}, we see that the parameter $\lambda$ which controls the relative importance of the low-rank term $L$ vs. the sparse term $S$ was set as $\frac{1}{\sqrt{\max \{m,n\}}}$, where $m$ and $n$ represent the numbers of rows and columns in the observation data matrix. To understand the relationships between the $L$, $S$ with the different values of the parameter $\lambda$, we chose the $A\beta+$ AD and $A\beta-$ CU group as the research objects. We find that the $rank(V)$ and $\Vert S \Vert_{\Omega,1}$ tend to decrease with an increase of the $\lambda$, as shown in Fig.~\ref{fig:difflam}. Because the threshold of the sparse component constraint is increasing with the increase of the $\lambda$ in the Eq.~\ref{eqn:resolveS}, the immediate result is the decrease of the $\Vert S \Vert_{\Omega,1}$. In addition, although the change of $\lambda$ is not directly related to the change of $rank(V)$, results indicate that the $rank(V)$ decrease due to gradually increasing removal of the noise and the sparse term $S$ with the increase of $\lambda$. And the $rank(V)$ is stable when $\lambda$ was greater than 0.006. However, when the $\lambda$ become too large, it results in homogeneous structures, reducing the discriminatory power for classifying the $A\beta+$ AD and $A\beta-$ CU groups. To validate the distinguishing power of the UMIs generated under different $\lambda$, we randomly use 90$\%$ of the $A\beta+$ AD and $A\beta-$ CU group as the training data to extract the common structures of these two groups with each $\lambda$ . The generated UMIs from remaining 10$\%$ of these two groups are classified using ROC analysis. This process is repeated 10 times. The average classification error rates with the different $\lambda$ are shown in Table~\ref{tab:classifilambda}. The results show that both high or low $\lambda$ will lead to an increase in classification error rate. This indicates that a low $\lambda$ introduces excessive noise and residual information of individuals and a high $\lambda$ makes the common structures of the two groups tend to be consistent. In both cases, the generated ROI cannot correctly reflect the morphological changes caused by AD, resulting in a decrease in UMI classification results. Based on the above discussion, $\lambda$ is empirically set as $0.0082$ throughout this paper.

\begin{table}[t]
\begin{center}
\begin{tabular}{cccc}
\hline\hline
 $\lambda$  & UMI-AD & UMI-CU	&  classification error  \\
\hline\hline
 0.003   & $11.57\pm6.09$ & $0.30\pm7.35$ &$0.1833$ \\
 0.004   & $11.17\pm6.26$ & $0.3364\pm7.18$ &$0.1839$ \\
 0.005   & $10.17\pm6.07$ & $0.7550\pm7.54$ &$0.1667$ \\
 0.006   & $13.08\pm5.95$ & $0.28\pm7.83$ &$0.1439$ \\
 0.007   & $13.77\pm5.87$ & $0.59\pm7.97$ &$0.1311$ \\
 0.008   & $13.87\pm4.08$ & $0.59\pm6.50$ &$0.1191$ \\
 0.009   & $13.91\pm4.12$ & $0.90\pm6.34$ &$0.1194$ \\
 0.01    & $13.37\pm5.36$ & $0.95\pm6.79$ &$0.1244$ \\
 0.011   & $12.61\pm6.26$ & $0.99\pm7.09$ &$0.1451$ \\
 0.012   & $12.54\pm5.97$ & $0.76\pm7.65$ &$0.1661$ \\
 0.013   & $10.79\pm6.32$ & $0.02\pm7.10$ &$0.1900$ \\
 0.014   & $11.36\pm6.77$ & $0.68\pm7.06$ &$0.1933$ \\
\hline\\
\end{tabular}
\caption{The classification error rates for the UMIs of $A\beta+$ AD and $A\beta-$ CU group with the different $\lambda$. The values of UMI mean are followed by standard deviations}\label{tab:classifilambda}
\end{center}
\end{table}

\subsection{Effect Size Comparison Between A$\beta+$ and A$\beta-$ subjects in the Longitudinal Study}\label{sec:EffSize}

To further validate the subspace decomposition-based UMIs have strong discrimination power for detecting the morphological changes induced by ADD, we analyze additional longitudinal A$\beta-$ subjects from ADNI database comparable to the A$\beta+$ group, which includes $32$ AD, $54$ MCI and $74$ CTL subjects. The demographic characteristic statistics information and the obtained UMIs for these $A\beta-$ samples are shown in Table~\ref{tab:demographic2}.
Similar to A$\beta+$ data, after the registration, surface deformation statistics are computed to obtain the raw RDs of all the longitudinal A$\beta-$ subjects. Based on the predefined ROIs (Fig.~\ref{fig:ROI} (a) and (c) in Section~\ref{sec:ROIextract}) and the AD atrophy degrees, we compute the UMIs of different longitudinal A$\beta-$ groups.

\begin{table}[t]
\begin{center}
\begin{tabular}{cccccc}
\hline\hline
& Gender(M/F) 	& Education   & Age    & MMSE & UMI\\
\hline\hline
AD baseline& \multirow{2}*{$18/14$}	& \multirow{2}*{$15.86\pm2.53$} & $76.56\pm6.34$  & $22.86\pm1.93$ & $11.97\pm7.15$ \\
AD m24& 	&  & $78.71\pm6.56$  & $18.87\pm3.78$ & $15.62\pm7.01$\\
\hline
MCI baseline& \multirow{2}*{$25/29$}	& \multirow{2}*{$15.61\pm2.44$} & $69.72\pm8.21$  & $28.48\pm1.62$ & $5.74\pm8.93$\\
MCI m24& 	&    & $71.86\pm8.21$  & $27.01\pm2.05$ & $8.77\pm8.79$\\
\hline
CU baseline& \multirow{2}*{$32/42 $} & \multirow{2}*{$15.89\pm2.60$} & $72.96\pm6.08$  & $29.15\pm1.13$ & $1.45\pm8.03$\\
CU m24&  &  & $75.06\pm6.04$  & $28.66\pm1.15$ & $4.36\pm7.59$\\
\hline
\end{tabular}
\caption{Demographic information of $A\beta-$ subjects in three clinical groups.}\label{tab:demographic2}
\end{center}
\end{table}

\begin{table}[b]
\begin{center}
\begin{tabular}{ccc}
\hline\hline
& UMIs 	& Volume Measures   \\
\hline\hline
AD &  $0.98$  & $0.72$ \\

\hline
MCI &  $0.73$  & $0.41$ \\
\hline
CTL &  $0.52$  & $0.27$ \\
\hline
\end{tabular}
\caption{The effect sizes from the differences between univariate biomarker measure (UMI and volume measure) changes of A$\beta+$ group and those of A$\beta-$ group in the same stage (AD, MCI, CU).}\label{tab:effectsize}
\end{center}
\end{table}

Then we use Cohen's d as effect size measure to test the differences between the univariate biomarkers (UMI vs. volume measure) changes of the longitudinal A$\beta+$ group and those of A$\beta-$ group in same clinical groups (AD, MCI and CU). The effect size comparison results are shown in Table~\ref{tab:effectsize}. The results show that the differences between UMI changes of the longitudinal A$\beta+$ group and those of the longitudinal A$\beta-$ group are more significant than the differences between volume changes of the longitudinal A$\beta+$ group and those of the longitudinal A$\beta-$ group. Since the UMIs are defined on the ROIs induced by ADD, it leads to larger changes in UMIs of longitudinal A$\beta+$ group, and smaller changes in UMIs of longitudinal A$\beta-$ group. The results  may also justify that the subspace decomposition-based UMIs have strong discrimination power for detecting the morphological changes induced by ADD.

\begin{table}
\begin{center}
\begin{tabular}{ccccc}
\hline\hline
  Group &  variables &R &$95\%$ CI of R  & $p$-value of R \\
\hline\hline
 \multirow{3}*{AD} & UMI vs. CDR-SB  & $0.44$ & $[0.26 \quad 0.59]$ & $9.26e-06$ \\
                   & UMI vs. ADAS-Cog11 & $0.49$ & $[0.32 \quad 0.63]$ & $6.74e-07$ \\
                   & UMI vs. MMSE   & $-0.50$ & $[-0.64 \quad -0.33]$ & $3.41e-07$ \\
 \hline
 \multirow{3}*{MCI} & UMI vs. CDR-SB  & $0.40$ & $[0.14 \quad 0.61]$ & $3.35e-03$ \\
                   & UMI vs. ADAS-Cog11 & $0.52$ & $[0.28 \quad 0.69]$ & $8.79e-05$ \\
                   & UMI vs. MMSE   & $-0.42$ & $[-0.62 \quad -0.16]$ & $2.16e-03$ \\
 \hline
 \multirow{3}*{CU} & UMI vs. CDR-SB  & $0.19$ & $[-0.06 \quad 0.41]$ & $0.14$ \\
                   & UMI vs. ADAS-Cog11 & $0.35$ & $[0.11 \quad 0.55]$ & $5.15e-03$ \\
                   & UMI vs. MMSE   & $-0.05$ & $[-0.29 \quad 0.19]$ & $0.68$ \\

\hline\\

\end{tabular}
\caption{The correlation analysis results between UMI and CDR-SB, ADAS-Cog11 and MMSE for different $A\beta+$ longitudinal groups.}\label{tab:relationship}
\end{center}
\end{table}

\subsection{Correlation Analysis with Cognitive Outcomes in Different Clinical Groups}\label{sec:CA}

In this section, we investigate whether changes of UMI results are correlated with changes of clinical rating scores within 24 months, such as Clinical Dementia Rating Sum of Boxes CDR-SB~\citep{Morris:pmid8232972}, AD Assessment Scale-Cognitive Subscale (ADAS-Cog11)~\citep{Rosen:AMJ1984} and MMSE scores. CDR-SB, ADAS-Cog11 scores are used to help track the progression of cognitive impairment, where a higher score indicates more severe dementia. Here we use the same $A\beta+$ longitudinal subjects ($46$ AD, $55$ MCI and $64$ CU in Section~\ref{sec:SPLD}) and with the Pearson parametric test~\citep{Lemasson:radio12}, we explore the correlation between the UMI changes and the clinical outcome changes within 24 months, including CDR-SB, ADAS-Cog11 and MMSE scores. The correlation analysis results, i.e., correlation coefficients (R), $95\%$ confidential intervals of R and correlation significance of $t$-test for R between the UMI changes and the changes of CDR-SB, ADAS-Cog11 and MMSE scores for different $A\beta+$ longitudinal groups are shown in Table~\ref{tab:relationship}.

Results show that the UMIs of $A\beta+$ longitudinal AD and MCI groups have moderate correlations with the CDR-SB, ADAS-Cog11 and MMSE scores. This is likely due to that the proposed UMIs well correlate with clinical outcomes within $A\beta+$ longitudinal AD and MCI groups. However, the correlations between the UMIs and the CDR-SB and MMSE scores of $A\beta+$ longitudinal CU group are not so significant as that of $A\beta+$ AD and MCI groups. 
From the perspective of data distribution, it is due to the fact that the distribution trend of the morphological changes depicted by UMIs differs from the distribution trend of the cognitive changes depicted by CDR-SB and MMSE scores within $A\beta+$ longitudinal CU group. A plausible reason may be that the morphological changes caused by AD occurred before the cognitive decline in the stage of unimpaired cognition, as proposed in the hypothetical model of AD biomarkers~\citep{Jack:Lancet13} and validated in a number of prior work~\citep{Sperling:PreAD11,Jack:ATN18,Weston:Neurology16,Zhao:NIMG17}, including our own study~\citep{Dong:NICL19}.

\begin{figure}[htp]
\begin{center}
\includegraphics[width=0.85\textwidth]{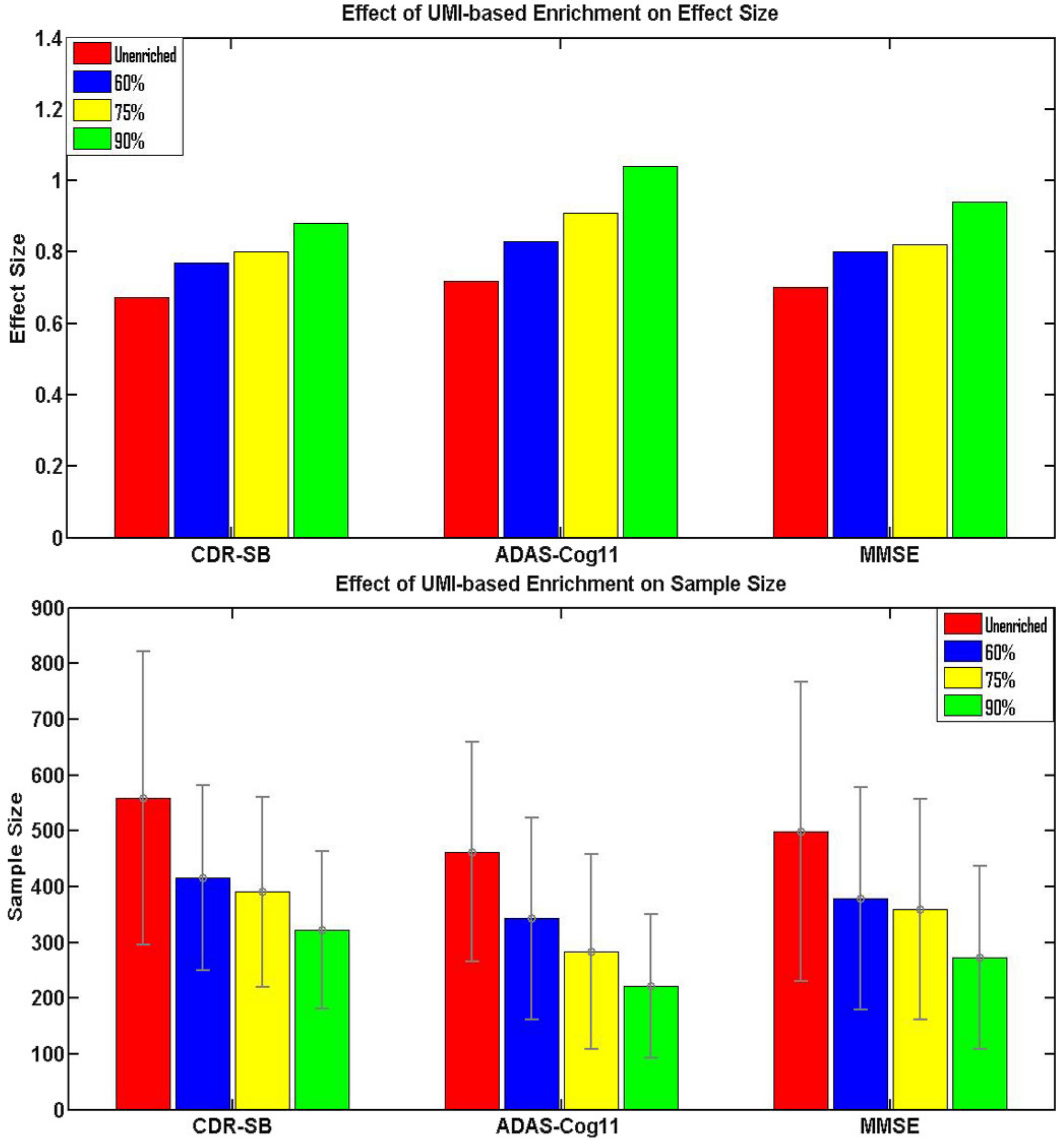}
\caption{Effect of UMI-based enrichment on effect sizes and sample sizes, as a function of cut-point ($60th$, $75th$ and $90th$ percentiles of the distribution of UMIs in the $A\beta-$ CU cohort) for CDR-SB, ADAS-Cog11 and MMSE measures. Error bars indicate $95\%$ confidence intervals of the estimated sample sizes.}
\label{fig:effect_sample_size}
\end{center}
\end{figure}

\subsection{Assessment of Enrichment Performance}\label{sec:enrichment}

To evaluate the enrichment performance of UMIs, we use a cut-point based enrichment strategy~\citep{Yu:pmid24211008} in $A\beta+$ longitudinal MCI group (the same MCI group in Section~\ref{sec:SPLD} denoted as unenriched group) to select the MCI subgroup (enriched group) and calculate the three clinical characteristics (CDR-SB, MMSE and ADAS-Cog11) of the selected MCI subgroup. The enrichment strategy is implemented as follows. First, a range of cutoff values are determined as the $60th$ , $75th$ and $90th$ percentiles of UMIs in the $A\beta-$ CU cohort ($271$ $A\beta-$ CU subjects in Section~\ref{sec:ROIextract}), which serve as a normative reference population. Secondly, for each cutoff value, we extract the clinical characteristics of the selected MCI subgroup and derive implications for clinical trials where only this subgroup will be enrolled. We calculate the effect size of the 24-months change in each of the three clinical measures. Denoting the effect size of the enriched group by $ES'$ and the unenriched group by $ES$, we then calculate the sample size $N'=(ES/ES')^2N$ required to achieve a statistical power equivalent to an unenriched group of size $N$~\citep{Yu:pmid24211008}. The comparison results of effect sizes and sample sizes of different clinical measures for the specific cut-points, i.e., $60th$ , $75th$ and $90th$ percentiles of the UMIs distribution of CU cohort, are shown in Fig.~\ref{fig:effect_sample_size}.

In the upper plot of Fig.~\ref{fig:effect_sample_size}, the results show that the effect sizes of all three clinical measures increase in all enriched group according to the specific cutoff values, comparing to the unenriched group. In general, more stringent cutoff values (higher UMI thresholds) lead to greater effect size. Cut-points corresponding to the $60th$, $75th$ and $90th$ percentiles of the $A\beta-$ CU cohort distribution yield effect sizes of $14.9\%-31.3\%$ (CDR-SB), $15.3\%-44.4\%$ (ADAS-Cog11) and $14.3\%-34.3\%$ (MMSE) greater than the unenriched group. The sample size results are shown in the lower plot of Fig.~\ref{fig:effect_sample_size}. Reduced sample sizes are obtained across the three cut-points for the three clinical measures. Error bars indicate $95\%$ confidence intervals of the estimated sample sizes. For cut-points corresponding to $60th$, $75th$ and $90th$ percentiles of the UMIs distribution of CU cohort, sample sizes corresponding to $57.8\%-74.5\%$ (CDR-SB), $47.8\%-74.0\%$ (ADAS-Cog11) and $54.6\%-75.9\%$ (MMSE) of the unenriched case ($100\%$) are estimated. This indicates that subspace decomposition-based UMIs provides an enrichment biomarker to select a fraction of the screening group that is likely to imminently progress most rapidly.

\subsection{Comparison with Heat Kernel Smoothing}\label{sec:RDUMIs}
\begin{figure}[t]
\begin{center}
\includegraphics[width=0.8\textwidth]{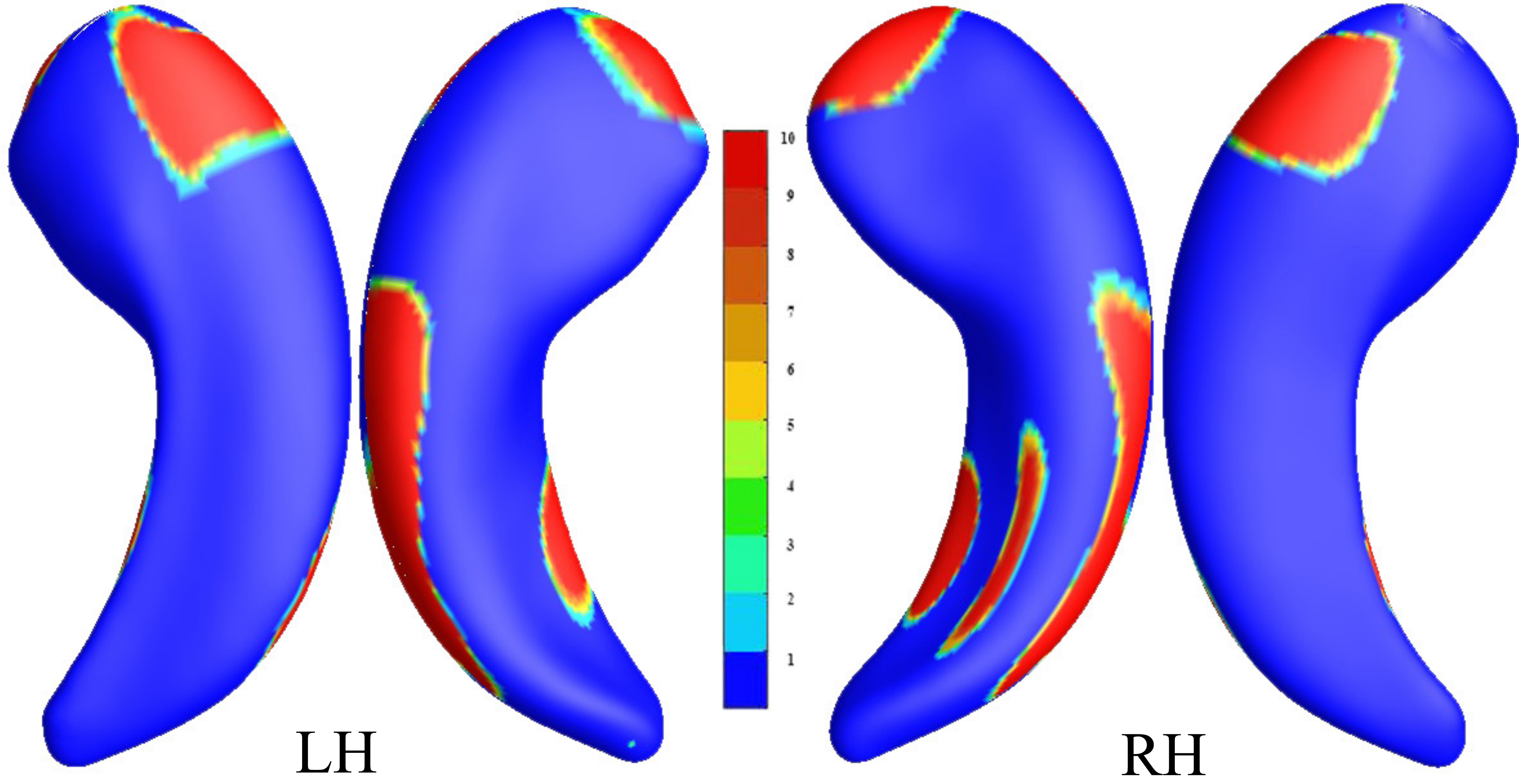}

\caption{The extracted ROI results for the $A\beta+$ AD and $A\beta-$ CU group differences based on the smoothed RD measures. All statistical ROI results were obtained by the two-sample $t$-test (corrected p-value$<0.0001$). Non-blue colors indicate the number of times a vertex is selected.}
\label{fig:ROI_smoothed}
\end{center}
\end{figure}

The motivation of our subspace decomposition work is to increase the signal-to-noise (SNR) and improve the sensitivity of statistical analysis. Surface smoothing work, especially heat kernel smoothing algorithm~\citep{Chung:nimg:2005}, has been adopted in our prior work~\citep[e.g.][]{Shi:NIMG17,Paquette:NICL17} to increase the SNR of surface tensor-based morphometry (TBM) features and boost the sensitivity of statistical analysis. Here we use the heat kernel smoothing results to generate UMIs and compare its performance with the one of our subspace decomposition-based UMIs. First, we use heat kernel smoothing (the kernel width=1 and the number of iteration=100) to generate the smoothed hippocampal RD measures of the same subjects in Section~\ref{sec:ROIextract}, i.e., $151$ $A\beta+$ AD patients and $271$ $A\beta-$ CU subjects. Second, following the content of Section 3.1, we randomly select $90\%$ of the subjects in each group to generate ROIs. This process is repeated 10 times. At each time, the regions of significant smoothed RD difference (ROIs) between $A\beta+$ AD group and $A\beta-$ CU group are assessed via the two-sample $t$-test with random field theory based $p$-value correction. Here, we use $p$-value correction method~\citep{Benjamini:FDR} to controlling the false discovery rate (FDR). The desired false discovery rate ($q$) is set as $0.0001$ and the generated ROIs are the vertices whose $t$-test corrected $p$-values represent the group differences smaller than $q$. The results of the generated ROIs are shown in Fig.~\ref{fig:ROI_smoothed}. Non-blue colors indicate the number of times a vertex is selected. The red color denotes that the vertex is selected for 10 times after 10 ROI generation processes. From the results, we can see that the selected ROIs based on the smoothed RD measures have a certain degree of similarity with the ROIs based on low-rank components. This indicated that heat kernel smoothing and subspace decomposition have similar effects in improving the stability of the ROI generation.

Using AD atrophy degrees and individual atrophy degrees on the generated ROIs based on the smoothed RD measures, we compute the UMIs of the longitudinal $A\beta+$ subjects according to the Eq.~\ref{eqn:MSSI}. With Eq.~\ref{eqn:n80}, we estimate the minimum sample sizes of UMIs based on the smoothed RD measures. The minimum sample sizes of the longitudinal $A\beta+$ AD, $A\beta+$ MCI and $A\beta+$ CU groups are 261, 374 and 546, respectively. The results show that the minimum sample sizes from the UMIs based on smoothed RD measures are all smaller than those based on raw RD measures. This indicates that heat kernel smoothing may increase the SNR and the sensitivity of statistical analysis. On the other hand, the results show that the minimum sample sizes from the heat kernel smoothing-based UMIs are all larger than low-rank decomposition-based UMIs, indicating that the low-rank decomposition-based UMIs gain more statistical power than those of the heat kernel smoothing-based UMIs. Part of the reason may be due to the fact that the decomposed low rank and subspace better characterizes group level morphometry features.

\subsection{Limitations}\label{sec:Limintations}
The proposed low-rank and sparse subspace decomposition-based UMI represents our initial efforts to develop robust ROI-based UNB. Nonetheless, there are several limitations of this study to consider. First, a relatively small number of subjects are included as the research objects, e.g. the ROIs are extracted from $151$ $A\beta+$ AD patients and $271$ $A\beta-$ CU subjects, which are certainly not enough to fully characterize the general morphological change patterns caused by ADD. Consequently, the range of morphological characteristics of individuals identified by the computed UMIs are likely incomplete. Besides, we have not tested our UMI framework together with the discovered ROIs in a cohort other than ADNI. Even so, our current results demonstrate the proposed approach may improve statistical power in a variety of UNB applications. Effect size comparison results also show that this proposed UMI measure is more sensitive to detect the hippocampal morphological changes than hippocampal volume measure. Second, we choose the RD as the hippocampal morphological feature because RD has been applied in several subcortical studies and served as an ideal description of the hippocampal structural changes induced by ADD. However, tensor-based morphometry~\citep{Chung:TMI08} and multivariate TBM~\citep{wang:nimg11} may be sensitive to deformations such as rotation, dilation, and shear along the surface tangent direction, and theorefore the TBM or mTBM may effectively capture hippocampal structural alterations (e.g. atrophy and enlargement) in tensor fields. The ability to describe the morphological features of the hippocampus will be enhanced if we combine the TBM or mTBM measures and the RD measure. Third, in this study, we are using UMIs alone to predict rates of progression from MCI to probable AD without combining other types of biomarkers, such as amyloid/tau/neurodegeneration (A/T/N) measurement framework~\citep{Jack:ATN16}, fluorodeoxyglucose positron emission tomography (FDG-PET) measurements of the regional cerebral metabolic rate for glucose (CMRgl) decline~\citep{Reiman:Neu2010} as a marker of clinical AD progression, fibrillar amyloid-$\beta$ PET measurements using Pittsburgh Compound B (PiB)~\citep{Ikonomovic:brain08} and cerebrospinal fluid (CSF) amyloid $\beta_{1-42}$ levels~\citep{Fagan:Neuro07} as biomarkers of AD pathology. If we combine these biomarkers with UMIs and carry out multivariate Cox analyses, the correctness of survival probabilities would likely be improved.

\section{Conclusion}
In this paper, we present a univariate neurodegeneration biomarker generation framework based on subspace decomposition to correctly and effectively depict the morphological changes induced by ADD. Applying the matrix decomposition and the local sparse constraints on the RD observation matrices, we obtain the ROIs closely associated with ADD which are robust to image noise. We then generate the UMIs to improve the statistical power for \emph{in vivo} MRI morphological analyses. Our empirical results demonstrate the potential that the UMIs may capture the ADD-induced brain morphometry abnormalities of the longitudinal groups at high risk for AD. Based on the optimal classification criterion, the UMIs have strong discrimination power for predicting the conversion rates from MCI subjects (converter and non-converter) to AD. The proposed framework is general and may be generalized to process other 3D brain imaging data, such as PET imaging data~\citep{Chen:HCI11}. In future studies, we will also evaluate our method in other brain imaging cohorts, such as Arizona APOE cohort~\citep{Caselli:Neurology04,Caselli:NEJM09} and OASIS cohort~\citep{oasis:MedRxiv19}.

\section{Acknowledgement}

This work was partially supported by National Natural Science Foundation of China (61772253, 61771231, 61873117, 61872170, 61903172); NSFC Joint Fund with Zhejiang Integration of Informatization and Industrialization under Key Project (U1609218); Key Research and Development Program of Shandong Province (2019JZZY010125); Shandong Province Higher Educational Science and Technology Program (J17KA050); National Institute on Aging (RF1AG051710, R21AG065942, R01AG031581, P30AG19610); National Institute of Biomedical Imaging and Bioengineering (R01EB025032); and Arizona Alzheimer's Consortium.

Data collection and sharing for this project was funded by the Alzheimer's Disease Neuroimaging Initiative (ADNI) (National Institutes of Health Grant U01 AG024904) and DOD ADNI (Department of Defense award number W81XWH-12-2-0012). ADNI is funded by the National Institute on Aging, the National Institute of Biomedical Imaging and Bioengineering, and through generous contributions from the following: Alzheimer's Association; Alzheimer's Drug Discovery Foundation; Araclon Biotech; BioClinica, Inc.; Biogen Idec Inc.; Bristol-Myers Squibb Company; Eisai Inc.; Elan Pharmaceuticals, Inc.; Eli Lilly and Company; EuroImmun; F. Hoffmann-La Roche Ltd and its affiliated company Genentech, Inc.; Fujirebio; GE Healthcare; IXICO Ltd.; Janssen Alzheimer Immunotherapy Research \& Development, LLC.; Johnson \& Johnson Pharmaceutical Research \& Development LLC.; Medpace, Inc.; Merck \& Co., Inc.; Meso Scale Diagnostics, LLC.; NeuroRx Research; Neurotrack Technologies; Novartis Pharmaceuticals Corporation; Pfizer Inc.; Piramal Imaging; Servier; Synarc Inc.; and Takeda Pharmaceutical Company. The Canadian Institutes of Rev December 5, 2013 Health Research is providing funds to support ADNI clinical sites in Canada. Private sector contributions are facilitated by the Foundation for the National Institutes of Health (www.fnih.org). The grantee organization is the Northern California Institute for Research and Education, and the study is coordinated by the Alzheimer's Disease Cooperative Study at the University of California, San Diego. ADNI data are disseminated by the Laboratory for Neuro Imaging at the University of Southern California.

\bibliographystyle{model2-names}
\bibliography{heatkernel}

\end{document}